\documentclass[10pt,twocolumn,letterpaper]{article}

\usepackage[pagenumbers]{cvpr}   

\usepackage[T1]{fontenc}
\usepackage{multirow}
\usepackage{tabularx}
\usepackage{algorithm}
\usepackage{algpseudocode}
\usepackage{listings}
\newcommand{\yellowbox}{\textcolor{Goldenrod}{$\square$}}
\newcommand{\greenbox}{\textcolor{ForestGreen}{$\square$}}
\newcommand{\redcircle}{\textcolor{red}{$\bigcirc$}}
\newcommand{\bgicon}{\raisebox{-0.15em}{\includegraphics[height=0.9em]{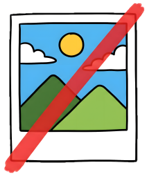}}}

\definecolor{cvprblue}{rgb}{0.21,0.49,0.74}
\usepackage[pagebackref,breaklinks,colorlinks,allcolors=cvprblue]{hyperref}

\def\confName{CVPR}
\def\confYear{2027}

\title{\textit{Choose What Returns}: Object-Aware KV Memory\\for Selective Recall in Interactive Video Generation}

\author{
Hyunwook Kang \qquad Junoh Park \qquad Hyunggyu Lim \qquad Taesup Kim\thanks{Corresponding author.}\\
Graduate School of Data Science\\ Seoul National University\\
}
\begin{document}
\maketitle
\begin{abstract}
Interactive long-video generation lets users build a story through successive prompts and reuse earlier characters and settings under new instructions.
Existing memory methods retain past content, yet they often fail at precise selective recall, returning the wrong appearance of a person, mixing different people, or keeping an unrequested background.
Based on our analysis of memory form and retrieval, we introduce an object-aware key--value (KV) memory framework for a frozen autoregressive diffusion generator.
It keeps the original KV of tracked object regions, selects background KV outside them by attention-scores, and links each memory with a readable record of its source scene and observed appearance.
An LLM selects the requested records, and a memory integrator supplies their KV for generation under the current prompt.
We also introduce a six-scene evaluation that tests selective attribute modification, time-specific appearance recall, and multi-object composition, in which each person's requested past appearance changes across requests.
Experiments show that our framework more reliably preserves each person's identity while reproducing their requested past appearance than existing memory methods, under both explicit scene references and natural-language cues.
\end{abstract}

\begin{figure*}[t]
\centering
\includegraphics[width=\textwidth]{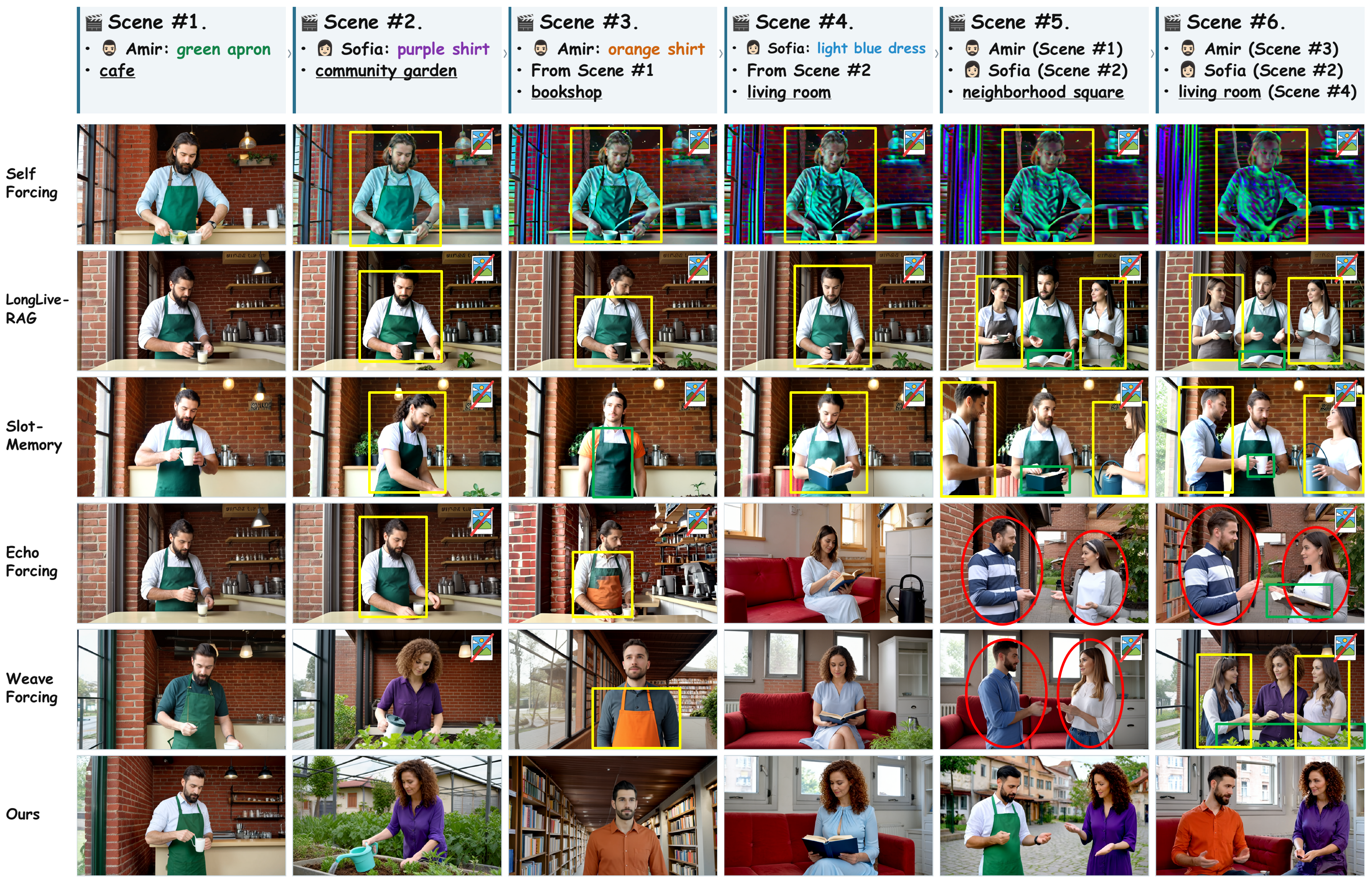}
\caption{\textbf{Qualitative comparison across six sequential scenes.} \textit{Natural} queries describe the source content without scene numbers. Yellow boxes (\yellowbox) mark prompt mismatches, red circles (\redcircle) inconsistencies with earlier scenes, green boxes (\greenbox) unwanted carryover, and \bgicon\ a mismatched background. In a simpler setting with more continuous background and fewer recalled elements, the same methods produce cleaner frames that better follow the requests, and these failures emerge once requests require more precise selective recall (Appendix~\ref{app:simple_setting}).}
\label{fig:motivation}
\end{figure*}

\begin{figure}[t]
\centering
\includegraphics[width=\linewidth]{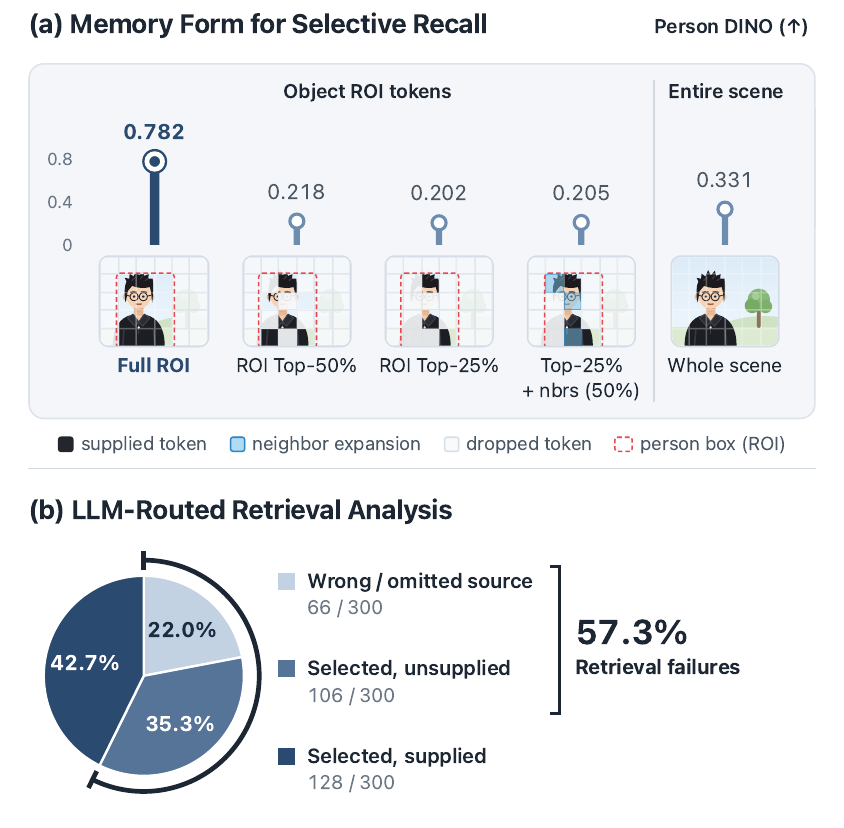}
\caption{\textbf{Memory form and retrieval matter for selective recall.} (a) DINO similarity~\cite{oquab2023dinov2} between source and generated person crops for each memory form. (b) Outcomes of 300 memory elements over 101 requests in a reimplementation of Weave Forcing. `Selected' means the correct source was chosen, and `unsupplied' means no KV position remained after masking.}
\label{fig:memory_analysis}
\end{figure}

\section{Introduction}
\label{sec:introduction}

Interactive long-video generation supports filmmaking and interactive storytelling by letting users build a video through successive prompts.
As the narrative develops, users may need to reuse earlier characters and settings under new instructions.
For example, a user may bring back an existing character with a new outfit (\textit{attribute modification}).
The user may also request the character's appearance from a particular earlier scene (\textit{time-specific recall}),
or combine characters and background from different scenes together (\textit{multi-object composition}).
These requests require precise control over which past elements return and what the current instruction changes.

Autoregressive diffusion models carry past visual information forward through cached key--value (KV) states~\cite{huang2025selfforcing,longlive2025}.
Memory extensions recover older content through scene recall and historical KV retrieval~\cite{wu2026echoforcing,longliverag2026}.
Finer-grained methods organize memory around objects~\cite{dou2026slotmemory} and use language-guided routing to select characters and backgrounds separately~\cite{weaveforcing2026}.
Several of these methods sparsely select which token KVs to store and read based on attention-scores, using techniques such as token compression and semantic masking.

However, access to past KV does not always ensure precise selective recall.
Figure~\ref{fig:motivation} shows unwanted character persistence, facial changes during attribute modification, and failures to reproduce characters and background from specific scenes.
These failures show that existing memory methods can fall short of precise selective recall, which current evaluations assuming largely unchanged characters and backgrounds may overlook (Appendix~\ref{app:simple_setting}).
Addressing these challenges requires examining both the form of visual memory supplied to the generator and the selection of relevant past observations.
This brings us to the fundamental question:
\textbf{How should visual memory be organized and retrieved for precise selective recall?}

We examine both aspects in our experiments. Figure~\ref{fig:memory_analysis}(a) shows that supplying object-region KV preserves appearance better than supplying the entire scene.
Within that region, full spatial coverage also yields higher appearance similarity than sparse attention-based token selection.
Figure~\ref{fig:memory_analysis}(b) reveals repeated retrieval failures, consisting of incorrect source assignments and selected-yet-unretrieved memories for which masking left no KV.
Together, these observations motivate separately selectable visual regions and a reliable link between the requested past observation and the KV memory supplied.

Guided by both findings, we introduce an object-aware KV memory framework for a frozen autoregressive diffusion generator.
It pairs spatially preserved visual memory with readable records that an LLM can identify and select.
Inspired by visual perception theories~\cite{egly1994shifting,qiu2007figureground}, we organize objects and backgrounds with distinguished retention rules.
A detector localizes objects, and a tracker follows their regions across frames.
We retain original KV within each tracked region, while background memory preserves attention-ranked KV outside detected objects.

Each record describes its source scene, object identifier, and observed appearance.
This metadata catalog makes the generated history understandable to an LLM, including how the same object appeared in different scenes.
An LLM router reasons over these descriptions and the user's request to select the relevant past observations.
A memory integrator gathers and combines their original KV.
Finally, the generator attends to this evidence KV alongside the ongoing scene's KV under the current prompt.

We evaluate selective attribute modification, time-specific appearance recall, and multi-object composition in six-scene histories, using both explicit scene references and natural-language source cues.
Each person's appearance changes during the history, and later requests ask for the appearance from a specific earlier scene, so repeating the most recent state does not succeed.
Under this evaluation, existing memory baselines often carry the most recent appearance forward, whereas our framework more often keeps each person's identity while reproducing their appearance from the requested scene.

Our contributions are threefold:
\begin{itemize}
    \item We analyze how object-level separation, spatial coverage, and historical source selection affect selective recall, and use these findings to guide memory formation and retrieval.
    \item We develop an object-aware KV framework that pairs spatially preserved object memories and separately compressed background memories with readable scene-specific records to leverage LLM reasoning and selection.
    \item We introduce a six-scene evaluation in which the requested past appearance of each person changes across requests, with separate measurements that directly test which appearance returns and expose failures that a single similarity score would hide.
\end{itemize}

\ifdefined\section
  \let\RWPreviewEnd\relax
\else
  \documentclass[10pt,twocolumn]{article}
  \usepackage[numbers,sort&compress]{natbib}
  \newcommand{\RWPreviewEnd}{%
    \bibliographystyle{plainnat}%
    \bibliography{references}%
    \end{document}%
  }
  \begin{document}
\fi

\section{Related Work}
\label{sec:related_work}

\paragraph{Interactive video generation.}
Autoregressive video diffusion generates video in chunks conditioned on cached KV states~\cite{causvid2025,huang2025selfforcing}.
LongLive refreshes the cache after prompt changes~\cite{longlive2025}.
Echo-Forcing stores scene memories for cross-scene recall~\cite{wu2026echoforcing}, while LongLive-RAG retrieves historical latents beyond the local context~\cite{longliverag2026}.
We examine how object-level separation, spatial coverage, and historical source selection affect selective recall.

\paragraph{Visual memory and retrieval.}
SlotMemory uses learned object slots to address original KV~\cite{dou2026slotmemory}.
IAMFlow maintains an identity registry with LLM-assigned global IDs and VLM-verified observations~\cite{liu2026advancingnarrativelongvideo}, while EM-Vid retrieves entity-indexed VAE latent patches through a structured script~\cite{vandersanden2026emvid}.
TetherMem gives subject and scene queries different access to history according to key region and memory age~\cite{li2026tethermem}.
Weave Forcing uses an LLM to route character and background slots to earlier shots, then masks unrelated content in compressed shot KV~\cite{weaveforcing2026}.
We instead retain original KV throughout tracked object regions localized by LLM-guided open-vocabulary detection, with background KV selected separately outside them.
Readable scene-specific records let an LLM select past observations from brief source cues.

\paragraph{Evaluation of video generation.}
VBench and T2V-CompBench evaluate general video quality and compositional generation~\cite{huang2023vbench,sun2025t2v}.
Memory evaluations emphasize continuity, scene return, and cross-shot composition~\cite{liu2026advancingnarrativelongvideo,dou2026slotmemory,wu2026echoforcing,weaveforcing2026}.
EM-Vid demonstrates attribute changes qualitatively~\cite{vandersanden2026emvid}, while VMem-Bench distinguishes historical fidelity from current-story correctness~\cite{chen2026vmembench}.
Our six-scene histories test whether each person's requested past appearance returns after attribute changes.
We measure selected-content fidelity, instruction following, and unwanted carryover separately.
\RWPreviewEnd

\begin{figure*}[t]
\centering
\includegraphics[width=\textwidth]{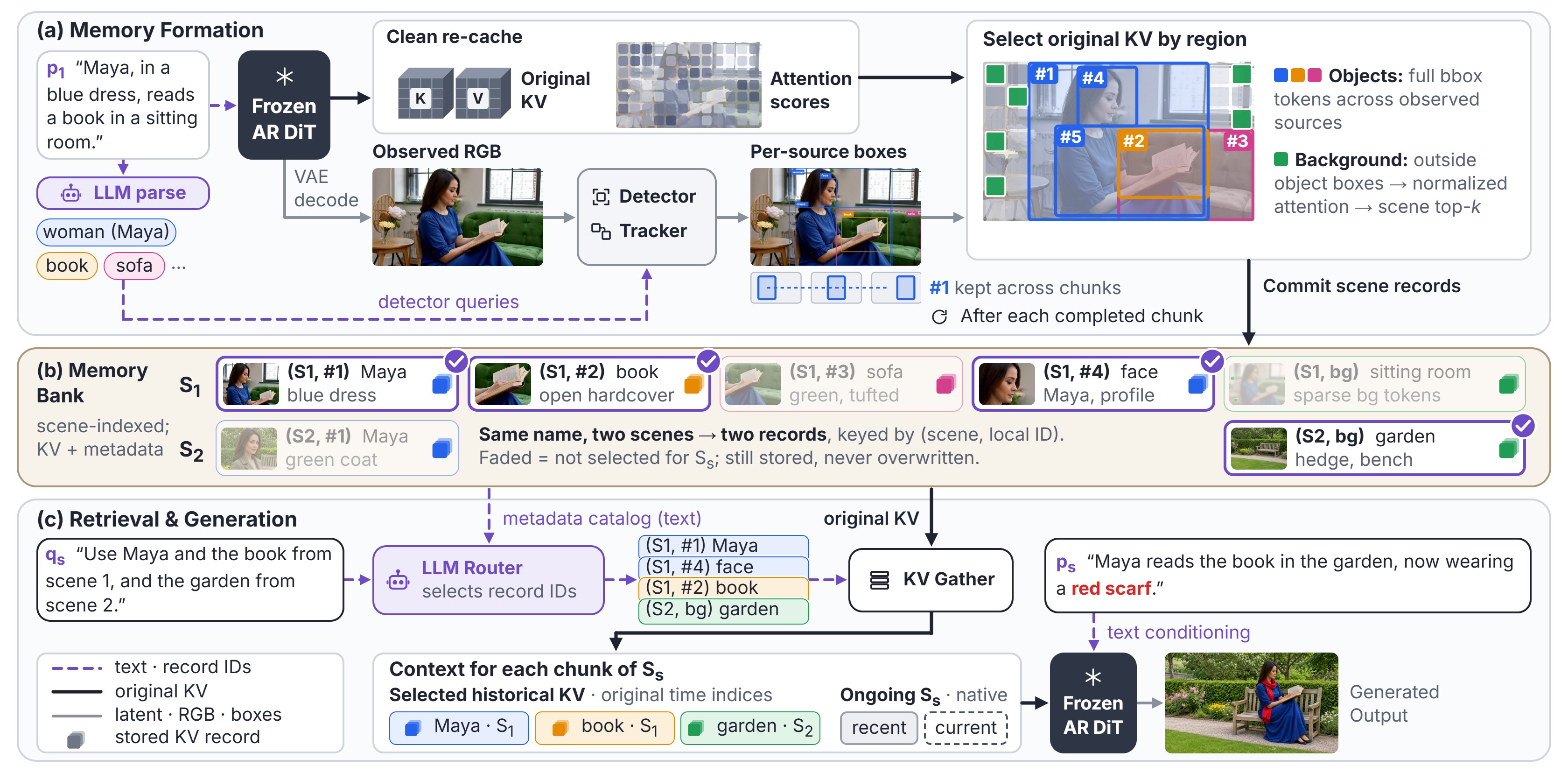}
\caption{\textbf{Overview of our framework.} (a) Memory formation keeps the full-box KV of tracked objects and selects sparse background KV by attention. (b) Scene records pair this KV with readable metadata, so a new look never overwrites an earlier one. (c) An LLM selects records from the catalog, and the generator attends to their KV together with the ongoing scene's KV under the current prompt. Frames and boxes come from our runs, whereas grids and the tracking sketch are schematic.}
\label{fig:method}
\end{figure*}

\section{Problem Setup}
\label{sec:problem_formulation}
\paragraph{Interactive generation.}

We consider text-to-video generation with a frozen autoregressive diffusion model. At scene index $s$, the model generates $X_s$ as a sequence of latent chunks. The content prompt $p_s$ describes the scene, and the reuse query $q_s$ specifies which past objects or backgrounds to recall. These inputs are provided separately or combined, depending on the method. The backbone uses cached keys and values to attend to previously generated frames. We retain these KV states as visual memory and supply selected states when generating later scenes.

\paragraph{Source-specific requests.}
A reuse request specifies which past person or background to use for each role in the current scene. We denote each role by $\ell$, such as the first person, second person, or background. Let $r<s$ index an earlier scene used as a source, and let $j$ identify an object or background within it. The requested assignment $\mathcal A_s^\star(\ell)=(r,j)$ specifies which past element should fill each role. The source scene $r$ also determines its appearance, which may differ across scenes. The query $q_s$ specifies the request through a scene reference or a description of a past location or activity. The retriever infers the assignment from $q_s$ and its available memory, while $\mathcal A_s^\star$ serves as the ground truth for evaluation.

\paragraph{Capabilities.}
We distinguish three capabilities for selective recall.
\emph{Selective attribute modification} changes requested attributes while preserving the person's identity and designated unchanged attributes.
\emph{Time-specific appearance recall} reproduces the past element's appearance from the specific scene.
We call the appearance of an element in a particular scene a \emph{version} of that element. After an attribute modification, a person has at least two versions, and time-specific recall asks for one of them.
\emph{Multi-object composition} combines selected people and backgrounds, potentially from different scenes.

\paragraph{Requirements.}
All three capabilities share the same three requirements.
\emph{Selected-content fidelity} requires each role to reproduce element $j$ as it appeared in source scene $r$, apart from the attributes that the current instruction changes.
\emph{Instruction following} requires the output to satisfy the current request, including requested changes and the requested place.
\emph{No unwanted carryover} requires content that was not requested, such as another version of a person or an extra person, to stay out of the scene.
Each requested person must appear as a separate person in the generated scene.

\paragraph{Memory and generation.}
Let $\mathcal M_{s-1}$ contain historical KV and its indexing information, such as source identifiers and descriptions.
Initially, $\mathcal M_0=\varnothing$.
We write scene-level memory selection, generation, and update as
\begin{equation}
\begin{aligned}
\mathcal K_s&=\mathcal R(p_s,q_s,\mathcal M_{s-1}),\\
(X_s,\mathcal Z_s)&=\mathcal G_\theta(p_s,q_s,\mathcal K_s;\xi_s),\\
\mathcal M_s&=\mathcal U(\mathcal M_{s-1},X_s,\mathcal Z_s,p_s,q_s).
\end{aligned}
\label{eq:scene_generation}
\end{equation}
The retrieval operator $\mathcal R$ selects and assembles historical keys and values into $\mathcal K_s$.
The generation procedure $\mathcal G_\theta$ produces scene $X_s$ using frozen generator parameters $\theta$ and sampling randomness $\xi_s$.
It also exposes a state stream $\mathcal Z_s$ for memory formation, including newly produced KV and, when used, attention scores.
The update operator $\mathcal U$ uses the KV states and any attention scores in $\mathcal Z_s$, the generated scene, and the current request to update the memory to $\mathcal M_s$. Memory updates can occur after each chunk is completed.

The selected memory $\mathcal K_s$ contains only KV from earlier scenes, while KV from the current scene is maintained separately during generation.
Memory updates use only observations and instructions available at that time.

\paragraph{Evaluation references.}
Each method is evaluated on its own generated history, whose source scenes may differ from their prompts, for example in shirt color.
We therefore measure selected-content fidelity against element $j$ as it appears in the method's own scene $r$, where $(r,j)=\mathcal A_s^\star(\ell)$, and instruction following against the written scenario requirements.

\section{Method}
\label{sec:method}

\subsection{Overview and Backbone}
\label{sec:method_workflow}

Figure~\ref{fig:method} summarizes our framework for memory formation, retrieval, and generation, and Algorithm~\ref{alg:framework} lists the steps for one scene (Appendix~\ref{app:procedures} details each part).
Memory formation organizes KV from generated scenes into object and background records indexed by source scene.
Retrieval selects these records for a later request.
The reuse query $q_s$ specifies which past content to recall, while the content prompt $p_s$ describes the current scene and requested attribute changes.
Retrieval precedes scene generation, while memory formation proceeds as the scene is generated.
\begin{algorithm}[t]
\caption{\textbf{One-scene generation in our framework.}}
\label{alg:framework}
\small
\begin{algorithmic}[1]
\Require prompt $p_s$, reuse query $q_s$, memory $\mathcal M_{s-1}$
\Ensure scene $X_s$, memory $\mathcal M_s$
\State $\widehat{\mathcal A}_s\gets\operatorname{LLM}\big(p_s,q_s,\operatorname{Catalog}(\mathcal M_{s-1})\big)$ \Comment{route}
\State $\mathcal K_s\gets\operatorname{IntegrateKV}(\mathcal M_{s-1},\widehat{\mathcal A}_s)$ \Comment{integrate}
\State $p_s^{+}\gets\operatorname{Refine}(p_s,\widehat{\mathcal A}_s)$ \Comment{refine prompt}
\For{each chunk $c$ of scene $s$} \Comment{generate and memorize}
  \State Denoise $c$ under $p_s^{+}$, attending to $\mathcal K_s$ \Comment{Eq.~\eqref{eq:controlled_memory_injection}}
  \State Detect and track objects, extend $\mathcal I^{\mathrm{ob}}_{s,j}$ \Comment{Eq.~\eqref{eq:object_memory_indices}}
  \State Update $\mathcal I^{\mathrm{bg}}_s$ by attention \Comment{Eq.~\eqref{eq:background_memory_indices}}
\EndFor
\State $\mathcal M_s\gets\mathcal M_{s-1}\cup\operatorname{Records}\big(\mathcal I^{\mathrm{ob}}_{s,\cdot},\mathcal I^{\mathrm{bg}}_s\big)$ \Comment{commit}
\end{algorithmic}
\end{algorithm}

We use a frozen autoregressive text-to-video diffusion transformer based on Self Forcing~\cite{huang2025selfforcing}.
It denoises one latent chunk at a time, conditioned on text and cached video KV.
After denoising, a clean forward pass recomputes KV for subsequent chunks.
We capture this KV and attention scores, and decode the completed chunk through the video VAE.
Detection operates on the decoded RGB.
The VAE's temporal mapping associates representative RGB frames with latent video timesteps, which we call sources.
We project pixel-space detection boxes onto the corresponding source's spatial token grid.

\subsection{Preserving Object Regions}
\label{sec:method_objects}
Visual perception studies suggest that objects and their boundaries guide how a scene is selectively processed~\cite{egly1994shifting,qiu2007figureground}.
Guided by this insight, we preserve object regions as spatial units and select background KV by attention score outside detected objects.
An LLM parses the current content prompt $p_s$ into object-category queries for an open-vocabulary detector, covering each person, their face, head hair, and garments, and the objects the prompt mentions.
The detector localizes objects in representative RGB frames.
A tracker associates observations within the scene and assigns scene-local instance IDs.
Only accepted observations contribute KV to an object's memory (Appendices~\ref{app:objects} and~\ref{app:tracking}).

Let $j$ denote a tracked object's scene-local ID in scene $s$.
The set $\mathcal V_{s,j}$ contains latent timesteps $t$ with accepted observations of that object.
The set $\Omega_{s,j,t}$ contains all token-grid cells overlapping its detected box at timestep $t$.
The object memory uses the index set
\begin{equation}
\mathcal I_{s,j}^{\mathrm{ob}}
=\{(t,v):t\in\mathcal V_{s,j},\ v\in\Omega_{s,j,t}\}.
\label{eq:object_memory_indices}
\end{equation}
Each pair $(t,v)$ identifies a token at latent timestep $t$ and spatial position $v$.
We gather the corresponding original keys and values from every transformer layer.
This retains every box-overlapping token at every accepted timestep in the scene.
The boxes are updated over time to follow object motion.

\subsection{Attention-Guided Background Selection}
\label{sec:method_background}

We store background KV separately so users can select a setting without selecting its original objects.
For completed chunk $c$ in scene $s$, let $\mathcal E_{s,c}$ contain the eligible background token indices.
A token is eligible if its grid cell overlaps no detected object box at the same latent timestep.
We use attention scores to rank these candidates.

Let $a_{c,i}$ denote the attention weight received by key $i$ in completed chunk $c$ during clean recaching.
We average over selected layers, all attention heads, and sampled queries from that chunk (Appendix~\ref{app:bg_scores}).
The attention softmax includes all video keys available in the clean pass, including historical memory when present.
Only current-chunk keys become new background candidates.
To compare relative importance across chunks, we divide each candidate's score by the mean over $\mathcal E_{s,c}$ and select across the scene:
\begin{equation}
\begin{aligned}
\widetilde a_{c,i}
 &=\frac{a_{c,i}}
 {|\mathcal E_{s,c}|^{-1}\sum_{v\in\mathcal E_{s,c}}a_{c,v}},\\
\mathcal I_s^{\mathrm{bg}}
 &=\operatorname{TopK}_{k_s}
 \left(\bigcup_{c\in s}\mathcal E_{s,c};\widetilde a\right).
\end{aligned}
\label{eq:background_memory_indices}
\end{equation}
The normalization applies to $i\in\mathcal E_{s,c}$.
The operator $\operatorname{TopK}_{k_s}$ returns up to $k_s$ token indices across all chunks in scene $s$, ranked by their normalized scores.
After each chunk, we merge its candidates with the retained set and keep the top $k_s$.
Each retained index supplies original K/V from every transformer layer.
We finalize the background record at scene completion.

\subsection{Scene Records, Routing, and Integration}
\label{sec:method_retrieval}

The memory integrator pairs each KV record with readable metadata: source scene, scene-local ID, memory type, name, observed appearance, and whether its KV is available for retrieval.
Names serve as user-facing aliases, and appearance descriptions summarize what was actually observed.
At scene completion, an LLM builds these records from the scene's detection lists alone.
It groups each person's accepted detections into face, head-hair, and garment components and writes each description from the accepted detector labels, so a description records what was detected rather than what the prompt requested.
Separate records preserve how the same named object appeared in different scenes, so later appearances do not overwrite earlier ones.

Given the reuse query $q_s$, an LLM reasons over the catalog to assign each requested role to a past record.
It selects scene-specific records rather than individual KV tokens:
\begin{equation}
\begin{aligned}
\widehat{\mathcal A}_s
  &=\operatorname{LLM}\!\left(p_s,q_s,\operatorname{Catalog}(\mathcal M_{s-1})\right),\\
\mathcal K_s
 &=\operatorname{IntegrateKV}(\mathcal M_{s-1},\widehat{\mathcal A}_s).
\end{aligned}
\label{eq:language_memory_routing}
\end{equation}
Here, $\operatorname{Catalog}$ returns the textual metadata, and $\widehat{\mathcal A}_s$ is the inferred assignment of requested roles to source records.
For each requested person, the router takes the face from the scene that introduced the person and the head hair and garments from the requested version, which anchors identity while the appearance follows the requested scene.
When the content prompt changes an attribute, the corresponding component is not recalled, so the prompt supplies the new attribute.
The router receives only text: the content prompt, the reuse query, and the catalog (Appendix~\ref{app:llm_components}).

The memory integrator implements $\operatorname{IntegrateKV}$ to convert these record selections into the historical KV context $\mathcal K_s$.
It checks record availability, merges the selected token-index sets, removes duplicate tokens (same source scene, latent timestep, and grid cell), and gathers the original K/V from every transformer layer (Appendix~\ref{app:integration}).
Unavailable selections are reported rather than replaced with unrelated memories, and the stored KV itself is never modified.

Recall prompts deliberately omit the appearance to be recalled, so the text condition alone does not describe the selected people.
We therefore insert the stored descriptions of the selected components into $p_s$ with a fixed template, producing the refined prompt $p_s^{+}$ that conditions generation.
Every instruction of $p_s$ is kept, and descriptions of components that $p_s$ changes are not inserted.

\subsection{Selected Memory Read and Generation}
\label{sec:method_injection}

At each scene boundary, we clear the previous scene's native KV cache.
The native cache then contains only recent and current-chunk KV from the ongoing scene.
Historical KV enters through the selected records.
Retrieved keys retain their original positional encoding, and their values remain unchanged.
Cache clearing does not reset the temporal indices used by the generator.

At each transformer layer, let $(K^n,V^n)$ denote the ongoing scene's recent clean KV and current noisy-chunk KV.
Let $(K^m,V^m)$ denote the selected historical KV from $\mathcal K_s$ at that layer.
For current-chunk queries $Q$, we use concatenated attention:
\begin{equation}
O=\operatorname{Attn}
\left(Q,[K^m;K^n],[V^m;V^n]\right).
\label{eq:controlled_memory_injection}
\end{equation}
Brackets denote concatenation along the token dimension.
Each generated chunk attends to this combination of selected history and ongoing-scene KV under the refined prompt $p_s^{+}$.
Memory-strength control, latent-amplitude correction, and decoder boundary handling are described in Appendices~\ref{app:mixing} and~\ref{app:latent}.
All generator parameters remain frozen.

\section{Experiments}
\label{sec:experiments}

We test whether a method brings back the people, appearances, and backgrounds requested from specific past scenes while following the current instruction.

\subsection{Scenarios and Comparison Setting}
\label{sec:experimental_setup}

\paragraph{Six-scene histories.}
Existing evaluations assess scene recall, cross-scene consistency, and memory-dependent state changes~\cite{wu2026echoforcing,weaveforcing2026,dou2026slotmemory,chen2026vmembench}.
Our scenarios target a stricter case, in which the same person appears in two versions and each request names the version to bring back.
Scenes~1--2 introduce two people in different places.
Scenes~3--4 test selective attribute modification by changing one person's clothing or hairstyle, so each person then has an initial and a modified version.
Scene~5 tests time-specific recall by bringing both people back in their initial versions, which are no longer the most recent ones.
Scene~6 tests multi-object composition with five branches that request different combinations of versions and earlier backgrounds.
In the notation of Section~\ref{sec:problem_formulation}, every scene specifies the requested assignment $\mathcal A_s^\star(\ell)=(r,j)$ (Appendix~\ref{app:benchmark_design}).

\paragraph{Queries and baselines.}
\textit{Explicit} queries name the source scene by number, and \textit{natural} queries refer to an earlier place or activity. Recall prompts never restate the clothing or hair to be recalled.
We adapt motifs from MemFlow~\cite{memflow2025} and evaluate ten scenarios.
We compare Self-Forcing, Echo-Forcing, LongLive-RAG, SlotMemory, Weave Forcing, and our method~\cite{huang2025selfforcing,wu2026echoforcing,longliverag2026,dou2026slotmemory,weaveforcing2026}. Weave Forcing and our method use the same LLM, \texttt{GPT-6.1 Sol high}~\cite{openai2026gpt61sol}, for routing.
A \emph{scenario} fixes two people, five places, and the six scene requests, and its five \emph{branches} differ only in the Scene~6 request. A \emph{history} is one six-scene video generated for one branch and query type, independently from Scene~1, so each method has 50 histories per query type.

\subsection{Evaluation Protocol}
\label{sec:experimental_evaluation}
\begin{table}[t]
\centering
\footnotesize
\setlength{\tabcolsep}{2pt}
\caption{\textbf{Metrics and the scenes where they apply.} Columns give each scene and the capability it tests: introduction (Int.), attribute modification (Mod.), time-specific recall (Rec.), and composition (Comp.). $^\ast$Reported in the appendix.}
\label{tab:metric_overview}
\begin{tabularx}{\linewidth}{@{}l >{\raggedright\arraybackslash}X cccc@{}}
\toprule
& & {\scriptsize Int.} & {\scriptsize Mod.} & {\scriptsize Rec.} & {\scriptsize Comp.} \\[-1pt]
Metric & Question & 1--2 & 3--4 & 5 & 6 \\
\midrule
\multicolumn{6}{@{}l}{\textit{Selected-content fidelity}} \\
Appear & Requested person, past appearance? & & & \checkmark & \checkmark \\
VersionPick & Closer to the requested version? & & & \checkmark & \checkmark \\
BG-G & Requested background reused? & & & & \checkmark \\
FaceKeep$^\ast$ & Identity kept during a change? & & \checkmark & & \\
\midrule
\multicolumn{6}{@{}l}{\textit{Instruction following}} \\
PromptAttr & Required attributes shown? & \checkmark & \checkmark & \checkmark & \checkmark \\
VersionAttr & Version-specific attributes correct? & & \checkmark & \checkmark & \checkmark \\
Place & Requested place shown? & \checkmark & \checkmark & \checkmark & \checkmark \\
Activity$^\ast$ & Requested activity shown? & \checkmark & \checkmark & \checkmark & \checkmark \\
\midrule
\multicolumn{6}{@{}l}{\textit{Unwanted carryover}} \\
Forbidden$^\ast$ & Excluded items absent? & \checkmark & \checkmark & \checkmark & \checkmark \\
Count & Number of people correct? & \checkmark & \checkmark & \checkmark & \checkmark \\
\midrule
\multicolumn{6}{@{}l}{\textit{VBench}} \\
Aesthetic & Visually appealing frames? & \checkmark & \checkmark & \checkmark & \checkmark \\
Imaging & Sharp, undistorted frames? & \checkmark & \checkmark & \checkmark & \checkmark \\
Text & Video matches the scene prompt? & \checkmark & \checkmark & \checkmark & \checkmark \\
\bottomrule
\end{tabularx}
\end{table}

\paragraph{What must be measured.}
The three requirements of Section~\ref{sec:problem_formulation} can fail independently: a method can follow the text but bring back a different person, or faithfully copy a source that never matched its own prompt.
We therefore measure each requirement with its own metrics (Table~\ref{tab:metric_overview}) and never combine them into one score.
Fidelity metrics compare the output with the method's own source scene, whereas instruction-following metrics compare it with the written requirements, so copying a wrong shirt color earns fidelity credit but no instruction-following credit.

\paragraph{Observations.}
We sample three frames per scene.
Faces are detected and embedded with SCRFD and a WebFace600K model~\cite{guo2022scrfd,deng2019arcface,zhu2021webface260m} and matched to the requested people one-to-one by face similarity~\cite{virtanen2020scipy}, so clothing never decides who is who.
Grounding DINO, SAM, and DINOv2 segment and encode torsos and backgrounds~\cite{liu2023groundingdino,kirillov2023segmentanything,oquab2023dinov2}.
A local Qwen2.5-VL-7B model reports visible attributes, prohibited items, the number of people, place, and activity without seeing method names or expected answers~\cite{bai2025qwen25vl}, and a separate step scores these reports against the requirements (Appendix~\ref{app:observer_v2}). Figure~\ref{fig:eval_pipeline} illustrates the full pipeline on one generated frame.

\paragraph{Fidelity of the requested person.}
Appear asks whether the requested person returns in the appearance from the requested scene.
Neither clothing nor face alone suffices, because another person can wear similar clothes and both versions of a person share one face.
For role $\ell$ with $\mathcal A_s^\star(\ell)=(r,j)$, we therefore combine three facial terms, based on person $j$'s first-scene face, with the torso similarity $C_\ell$ to person $j$ in scene $r$:
\begin{equation}
\mathrm{Appear}_\ell
=100\left(
F_\ell^{\mathrm{abs}}
F_\ell^{\mathrm{rel}}
F^{\mathrm{sep}}
C_\ell
\right)^{1/4}.
\label{eq:appearance_index_v2}
\end{equation}
$F_\ell^{\mathrm{abs}}$ rewards resemblance of the generated face to the intended person and $F_\ell^{\mathrm{rel}}$ an advantage over the other person, $F^{\mathrm{sep}}$ downweights histories whose two initial identity references are insufficiently distinct, and $C_\ell$ averages DINOv2 and color-histogram similarity (Appendix~\ref{app:appearance_v2}).
The geometric mean limits how much a strong torso match can compensate for weak facial similarity.
Appear applies to Scenes~5--6. In Scenes~3--4, where the clothing changes, FaceKeep checks identity alone.

\paragraph{Fidelity of the requested version.}
Because both versions share the face and often most of the clothing, Appear changes only partly when the wrong version returns.
VersionPick asks directly which version the output resembles by comparing the generated torso with the torsos of both versions:
\begin{equation}
\mathrm{VersionPick}_\ell=100\cdot\mathbf{1}\!\left[C_\ell^{\mathrm{req}}>C_\ell^{\mathrm{other}}\right].
\label{eq:version_pick}
\end{equation}
Averaged over frames, 50 is chance and lower values mean that the other version returned. VersionPick applies to Scenes~5--6 when the two versions differ in clothing.

\paragraph{Fidelity of the requested background.}
BG-G measures how much of the requested source background is reused in Scene~6, rewarding matched DINOv2 patches that keep a consistent layout after people are masked (Appendix~\ref{app:bgg}).

\begin{table*}[t]
\centering
\fontsize{8.5}{10}\selectfont
\setlength{\tabcolsep}{2.4pt}
\caption{\textbf{Selective-recall results.} Explicit / Natural queries on a 0--100 scale, averaged over 50 histories per method and query type with equal weight per scenario. Chance is 50 for VersionPick and about 20 for Place. Aesthetic, Imaging, and Text (text--video alignment) are VBench scores. Best in bold, second best underlined.}
\label{tab:main_v2}
\begin{tabular}{l cccccccccc}
\toprule
 &  &  &  &  &  &  & & \multicolumn{3}{c}{VBench} \\
\cmidrule(lr){9-11}
Method & Appear $\uparrow$ & PromptAttr $\uparrow$ & VersionAttr $\uparrow$ & VersionPick $\uparrow$ & Count $\uparrow$ & BG-G $\uparrow$ & Place $\uparrow$ & Aesthetic $\uparrow$ & Imaging $\uparrow$ & Text $\uparrow$ \\
\midrule
Self-Forcing & 21.3\,/\,21.1 & 33.0\,/\,27.1 & 10.2\,/\,6.2 & 30.0\,/\,29.6 & 83.3\,/\,83.3 & \textbf{64.9}\,/\,\textbf{63.3} & 20.2\,/\,19.7 & 61.5\,/\,61.7 & 71.6\,/\,70.2 & 8.5\,/\,8.6 \\
Echo-Forcing & 21.9\,/\,21.7 & 43.4\,/\,47.6 & 36.5\,/\,39.9 & 36.8\,/\,45.5 & \underline{99.2}\,/\,97.5 & 7.6\,/\,9.6 & 40.1\,/\,38.7 & \underline{64.3}\,/\,\textbf{65.9} & 74.4\,/\,\textbf{74.7} & 16.9\,/\,15.6 \\
LongLive-RAG & 37.2\,/\,37.5 & 35.1\,/\,37.1 & 18.2\,/\,20.6 & 38.6\,/\,31.5 & 93.0\,/\,93.7 & 57.0\,/\,\underline{60.3} & 18.3\,/\,20.7 & \textbf{65.3}\,/\,\underline{65.5} & 74.5\,/\,74.4 & 11.7\,/\,11.6 \\
SlotMemory & 37.7\,/\,39.3 & 47.5\,/\,45.6 & 36.2\,/\,36.0 & \underline{55.7}\,/\,41.9 & 94.1\,/\,93.4 & \underline{59.3}\,/\,53.5 & 29.5\,/\,25.0 & 63.7\,/\,64.0 & 74.5\,/\,74.3 & 15.9\,/\,14.2 \\
Weave Forcing & \underline{38.7}\,/\,\underline{40.1} & \underline{60.4}\,/\,\underline{60.0} & \underline{52.2}\,/\,\underline{41.8} & 47.9\,/\,\underline{50.2} & 97.6\,/\,\underline{97.9} & 25.7\,/\,34.2 & \underline{65.5}\,/\,\underline{61.0} & 63.7\,/\,64.5 & \underline{74.7}\,/\,\underline{74.5} & \underline{20.6}\,/\,\underline{19.7} \\
\textbf{Ours} & \textbf{60.2}\,/\,\textbf{62.6} & \textbf{93.7}\,/\,\textbf{91.6} & \textbf{91.6}\,/\,\textbf{88.6} & \textbf{94.2}\,/\,\textbf{93.0} & \textbf{99.5}\,/\,\textbf{99.6} & 39.0\,/\,27.6 & \textbf{88.7}\,/\,\textbf{88.5} & 63.9\,/\,63.6 & \textbf{74.9}\,/\,\underline{74.5} & \textbf{22.0}\,/\,\textbf{22.0} \\
\bottomrule
\end{tabular}
\end{table*}

\paragraph{Instruction following.}
PromptAttr checks the clothing and hair attributes required in the current scene, and VersionAttr checks only the attributes that differ between a person's two versions, using the value of the requested version (Appendix~\ref{app:attributes}).
Place checks whether the observer picks the requested place among all places of the scenario, so chance is about 20 (Appendix~\ref{app:place}).
Each attribute or place scores 100 when it matches, zero when it is clearly wrong, and stays unknown when it is not visible.

\paragraph{Unwanted carryover.}
Forbidden checks for objects or garments that the request or the requested appearance excludes.
Count gives partial credit $100\cdot\min(n,n^\star)/\max(n,n^\star)$ for $n$ observed and $n^\star$ requested people, using only counts on which differently worded questions agree (Appendix~\ref{app:missingness_v2}).

\paragraph{How the metrics complement each other.}
Appear rewards a faithful copy of a wrong source, which PromptAttr catches. PromptAttr counts unchanged attributes, which VersionAttr removes. VersionAttr depends on attribute labels, which VersionPick avoids by comparing the versions visually. BG-G rewards a reused but wrong room, which Place catches.
Count and Forbidden capture extra people and objects that one-to-one face matching ignores, so an extra person never erases the evidence about appearance.

\paragraph{Aggregation.}
Within a history, every scene where a metric applies receives equal weight. We then average repeated generations, branches, and scenarios with equal weight, separately for explicit and natural queries.
Unknown outcomes are left out of the mean, and their possible effect is reported as a range in Appendices~\ref{app:full_results_v2} and~\ref{app:robustness_v2}.
Its Aesthetic and Imaging scores stay within 2.3 points of the best baseline, and its Text score is the highest.

\subsection{Results}
\label{sec:experimental_analyses}
\begin{table*}[t]
\centering
\small
\setlength{\tabcolsep}{4pt}
\caption{\textbf{Ablation study on six-scene histories} (\textit{Explicit} and \textit{natural} queries pooled). Each variant replaces one component of the full method. Best in bold, second best underlined.}
\label{tab:ablation_v2}
\begin{tabular}{l cccc ccc cc}
\toprule
 & \multicolumn{4}{c}{Selected-content fidelity $\uparrow$} & \multicolumn{3}{c}{Instruction following $\uparrow$} & \multicolumn{2}{c}{Carryover $\uparrow$} \\
\cmidrule(lr){2-5}\cmidrule(lr){6-8}\cmidrule(lr){9-10}
Variant & Appear & VersionPick & FaceKeep & BG-G & PromptAttr & VersionAttr & Place & Exact count & Forbidden \\
\midrule
Full method & \textbf{64.3} & \textbf{93.8} & \textbf{43.5} & \textbf{10.8} & \underline{90.5} & \underline{87.5} & \textbf{100.0} & \textbf{100.0} & \textbf{100.0} \\
w.o.\ Prompt refinement & 50.3 & 66.7 & 28.3 & 0.0 & 75.3 & 65.6 & 74.2 & \textbf{100.0} & 50.0 \\
w.o.\ Object-level extraction & \underline{58.7} & \underline{83.3} & \underline{39.3} & \underline{7.8} & \textbf{91.3} & \textbf{93.8} & \underline{87.5} & 95.8 & \textbf{100.0} \\
w.o.\ LLM router (BM25) & 51.8 & 70.8 & 35.8 & 1.5 & 88.4 & 84.4 & 79.2 & \textbf{100.0} & \textbf{100.0} \\
\bottomrule
\end{tabular}
\end{table*}

\paragraph{Overall comparison.}
Our method leads on every recall metric except BG-G under both query types (Table~\ref{tab:main_v2}).
Under explicit queries, it reaches 60.2 Appear against 38.7 for the strongest baseline, 91.6 VersionAttr against 52.2, and 88.7 Place against 65.5. Under natural queries, it reaches 62.6 Appear and 93.0 VersionPick against at most 40.1 and 50.2.
Because Appear requires both the intended face and the torso from the requested scene, this lead shows that our method keeps each person's identity while reproducing their requested past appearance.
Its Aesthetic and Imaging scores stay within 2.3 points of the best baseline, and its Text score is the highest (22.0 under both query types), so these gains do not reduce frame quality or prompt alignment.

\paragraph{Returning to an older version.}
In Scene~5, where the most recent appearance is the wrong one, Self-Forcing never returns the initial version under explicit queries (VersionPick 0.0), and Echo-Forcing and LongLive-RAG rarely do (24.3 and 20.0), whereas our method reaches 93.7.
Across Scenes~5--6, every baseline stays between 29.6 and 55.7, close to or below chance: access to past KV does not by itself bring back the requested version.

\paragraph{What each measurement reveals.}
Self-Forcing obtains the highest BG-G (64.9 and 63.3), yet its VersionAttr is 10.2 and its Place stays at chance (20.2), because it keeps the same people, clothing, and room throughout the history.
Echo-Forcing reaches a nearly perfect Count (99.2) but one of the lowest Appear scores (21.9): it shows the requested number of people, but rarely the requested people.
One column alone would misjudge either method. Appendix~\ref{app:qualitative} illustrates these patterns, Appendix~\ref{app:full_results_v2} reports further metrics, and Appendix~\ref{app:simple_setting} contrasts a simpler setting.

\paragraph{Ablation study.}
Table~\ref{tab:ablation_v2} replaces one component at a time in six-scene histories (Appendix~\ref{app:ablation_v2}).
Without prompt refinement, VersionAttr drops from 87.5 to 65.6, Place from 100.0 to 74.2, and Forbidden from 100.0 to 50.0, showing that the stored descriptions keep the text consistent with the recalled version.
Whole-scene KV lowers Appear from 64.3 to 58.7 and VersionPick from 93.8 to 83.3 and adds extra people (exact count 95.8), although the attributes are kept.
Replacing the LLM router with BM25 lowers VersionPick to 70.8 and Place to 79.2, since BM25 selects different source scenes for some requests.

\section{Conclusion}
\label{sec:conclusion}

We presented an object-aware KV memory framework for selective recall in interactive video generation.
It keeps the original KV of tracked object regions, stores attention-selected background KV separately, and pairs each memory with a readable record that an LLM router uses to select what a request refers to.
In six-scene histories where the requested past appearance of each person changes across requests, our framework more reliably kept each person's identity while reproducing the requested past appearance than existing memory methods, under both explicit and natural-language queries.
These results indicate that precise selective recall depends on both the form of the stored visual memory and a reliable link between a request and the past observation it refers to.
Appendix~\ref{app:limitations} discusses the limitations.

{
    \small
    \bibliographystyle{ieeenat_fullname}
    \bibliography{main}

\begin{thebibliography}{37}
\providecommand{\natexlab}[1]{#1}
\providecommand{\url}[1]{\texttt{#1}}
\expandafter\ifx\csname urlstyle\endcsname\relax
  \providecommand{\doi}[1]{doi: #1}\else
  \providecommand{\doi}{doi: \begingroup \urlstyle{rm}\Url}\fi

\bibitem[Bai et~al.(2025)Bai, Chen, Liu, Wang, Ge, Song, Dang, Wang, Wang, Tang, Zhong, Zhu, Yang, Li, Wan, Wang, Ding, Fu, Xu, Ye, Zhang, Xie, Cheng, Zhang, Yang, Xu, and Lin]{bai2025qwen25vl}
Shuai Bai, Keqin Chen, Xuejing Liu, Jialin Wang, Wenbin Ge, Sibo Song, Kai Dang, Peng Wang, Shijie Wang, Jun Tang, Humen Zhong, Yuanzhi Zhu, Mingkun Yang, Zhaohai Li, Jianqiang Wan, Pengfei Wang, Wei Ding, Zheren Fu, Yiheng Xu, Jiabo Ye, Xi Zhang, Tianbao Xie, Zesen Cheng, Hang Zhang, Zhibo Yang, Haiyang Xu, and Junyang Lin.
\newblock {Qwen2.5-VL} technical report.
\newblock \emph{arXiv preprint arXiv:2502.13923}, 2025.

\bibitem[Barath et~al.(2020)Barath, Noskova, Ivashechkin, and Matas]{barath2020magsacplusplus}
Daniel Barath, Jana Noskova, Maksym Ivashechkin, and Jiri Matas.
\newblock {MAGSAC++}, a fast, reliable and accurate robust estimator.
\newblock In \emph{Proceedings of the IEEE/CVF Conference on Computer Vision and Pattern Recognition}, pages 1304--1312, 2020.

\bibitem[Chen et~al.(2026)Chen, Shi, Wang, Xu, Wang, Hong, Zhang, Ma, and Zhang]{chen2026vmembench}
Yuzhuo Chen, Huafeng Shi, Xinyu Wang, Yulong Xu, Yucheng Wang, Haoqin Hong, Guoxin Zhang, Zehua Ma, and Weiming Zhang.
\newblock More memory, better videos? a controlled study of visual memory in long-video generation.
\newblock \emph{arXiv preprint arXiv:2610.35001}, 2026.

\bibitem[Dao(2024)]{dao2024flashattention}
Tri Dao.
\newblock {FlashAttention-2}: Faster attention with better parallelism and work partitioning.
\newblock In \emph{International Conference on Learning Representations}, 2024.

\bibitem[Deng et~al.(2019)Deng, Guo, Xue, and Zafeiriou]{deng2019arcface}
Jiankang Deng, Jia Guo, Niannan Xue, and Stefanos Zafeiriou.
\newblock {ArcFace}: Additive angular margin loss for deep face recognition.
\newblock In \emph{Proceedings of the IEEE/CVF Conference on Computer Vision and Pattern Recognition}, pages 4690--4699, 2019.

\bibitem[Dou et~al.(2026)Dou, Li, Cui, Zhou, Wang, and Zhu]{dou2026slotmemory}
Weijia Dou, Hui Li, Jiahao Cui, Lei Zhou, Jingdong Wang, and Siyu Zhu.
\newblock {SlotMemory}: Object-centric {KV} memory for streaming long-video generation.
\newblock arXiv preprint arXiv:2605.31033, 2026.

\bibitem[Egly et~al.(1994)Egly, Driver, and Rafal]{egly1994shifting}
R. Egly, J. Driver, and R.~D. Rafal.
\newblock Shifting visual attention between objects and locations: Evidence from normal and parietal lesion subjects.
\newblock \emph{Journal of Experimental Psychology: General}, 123\penalty0 (2):\penalty0 161--177, 1994.

\bibitem[Guo et~al.(2022)Guo, Deng, Lattas, and Zafeiriou]{guo2022scrfd}
Jia Guo, Jiankang Deng, Alexandros Lattas, and Stefanos Zafeiriou.
\newblock Sample and computation redistribution for efficient face detection.
\newblock In \emph{International Conference on Learning Representations}, 2022.

\bibitem[Howard et~al.(2019)Howard, Sandler, Chu, Chen, Chen, Tan, Wang, Zhu, Pang, Vasudevan, Le, and Adam]{howard2019mobilenetv3}
Andrew Howard, Mark Sandler, Grace Chu, Liang-Chieh Chen, Bo Chen, Mingxing Tan, Weijun Wang, Yukun Zhu, Ruoming Pang, Vijay Vasudevan, Quoc~V. Le, and Hartwig Adam.
\newblock Searching for {MobileNetV3}.
\newblock In \emph{Proceedings of the IEEE/CVF International Conference on Computer Vision}, pages 1314--1324, 2019.

\bibitem[Hu et~al.(2026)Hu, Yang, Huang, Han, and Chen]{longliverag2026}
Qixin Hu, Shuai Yang, Wei Huang, Song Han, and Yukang Chen.
\newblock {LongLive-RAG}: A general retrieval-augmented framework for long video generation.
\newblock arXiv preprint arXiv:2606.02553, 2026.

\bibitem[Huang et~al.(2026)Huang, Li, He, Zhou, and Shechtman]{huang2025selfforcing}
Xun Huang, Zhengqi Li, Guande He, Mingyuan Zhou, and Eli Shechtman.
\newblock Self forcing: Bridging the train-test gap in autoregressive video diffusion.
\newblock \emph{Advances in Neural Information Processing Systems}, 38:\penalty0 167283--167308, 2026.

\bibitem[Huang et~al.(2024)Huang, He, Yu, Zhang, Si, Jiang, Zhang, Wu, Jin, Chanpaisit, Wang, Chen, Wang, Lin, Qiao, and Liu]{huang2023vbench}
Ziqi Huang, Yinan He, Jiashuo Yu, Fan Zhang, Chenyang Si, Yuming Jiang, Yuanhan Zhang, Tianxing Wu, Qingyang Jin, Nattapol Chanpaisit, Yaohui Wang, Xinyuan Chen, Limin Wang, Dahua Lin, Yu Qiao, and Ziwei Liu.
\newblock {VBench}: Comprehensive benchmark suite for video generative models.
\newblock In \emph{Proceedings of the IEEE/CVF Conference on Computer Vision and Pattern Recognition}, pages 21807--21818, 2024.

\bibitem[Ji et~al.(2025)Ji, Chen, Yang, Tao, Wan, and Zhao]{memflow2025}
Sihui Ji, Xi Chen, Shuai Yang, Xin Tao, Pengfei Wan, and Hengshuang Zhao.
\newblock {MemFlow}: Flowing adaptive memory for consistent and efficient long video narratives.
\newblock arXiv preprint arXiv:2512.14699, 2025.

\bibitem[Ke et~al.(2021)Ke, Wang, Wang, Milanfar, and Yang]{ke2021musiq}
Junjie Ke, Qifei Wang, Yilin Wang, Peyman Milanfar, and Feng Yang.
\newblock {MUSIQ}: Multi-scale image quality transformer.
\newblock In \emph{Proceedings of the IEEE/CVF International Conference on Computer Vision}, pages 5148--5157, 2021.

\bibitem[Kirillov et~al.(2023)Kirillov, Mintun, Ravi, Mao, Rolland, Gustafson, Xiao, Whitehead, Berg, Lo, Dollár, and Girshick]{kirillov2023segmentanything}
Alexander Kirillov, Eric Mintun, Nikhila Ravi, Hanzi Mao, Chloe Rolland, Laura Gustafson, Tete Xiao, Spencer Whitehead, Alexander~C. Berg, Wan-Yen Lo, Piotr Dollár, and Ross Girshick.
\newblock Segment anything.
\newblock In \emph{Proceedings of the IEEE/CVF International Conference on Computer Vision}, pages 4015--4026, 2023.

\bibitem[Li et~al.(2026)Li, Zhang, Zhou, Zuo, Wang, Zhou, Sang, and Gao]{li2026tethermem}
Chen Li, Peng Zhang, Hanyu Zhou, Jialong Zuo, Fei Wang, Daiguo Zhou, Nong Sang, and Changxin Gao.
\newblock Tether the subject, release the scene: Query-aware memory routing for long-horizon autoregressive video generation.
\newblock arXiv preprint arXiv:2608.26902, 2026.

\bibitem[Lin et~al.(2014)Lin, Maire, Belongie, Hays, Perona, Ramanan, Doll{\'a}r, and Zitnick]{lin2014coco}
Tsung-Yi Lin, Michael Maire, Serge Belongie, James Hays, Pietro Perona, Deva Ramanan, Piotr Doll{\'a}r, and C.~Lawrence Zitnick.
\newblock {Microsoft COCO}: Common objects in context.
\newblock In \emph{European Conference on Computer Vision}, pages 740--755, 2014.

\bibitem[Liu et~al.(2026)Liu, Zhang, Jiang, Wang, Liang, Xue, Yi, and Liu]{liu2026advancingnarrativelongvideo}
Jinzhuo Liu, Jiangning Zhang, Wencan Jiang, Yabiao Wang, Dingkang Liang, Zhucun Xue, Ran Yi, and Yong Liu.
\newblock Advancing narrative long video generation via training-free identity-aware memory.
\newblock In \emph{Advances in Neural Information Processing Systems}, 2026.

\bibitem[Liu et~al.(2024)Liu, Zeng, Ren, Li, Zhang, Yang, Jiang, Li, Yang, Su, Zhu, and Zhang]{liu2023groundingdino}
Shilong Liu, Zhaoyang Zeng, Tianhe Ren, Feng Li, Hao Zhang, Jie Yang, Qing Jiang, Chunyuan Li, Jianwei Yang, Hang Su, Jun Zhu, and Lei Zhang.
\newblock {Grounding DINO}: Marrying dino with grounded pre-training for open-set object detection.
\newblock In \emph{European Conference on Computer Vision}, 2024.

\bibitem[Liu et~al.(2016)Liu, Anguelov, Erhan, Szegedy, Reed, Fu, and Berg]{liu2016ssd}
Wei Liu, Dragomir Anguelov, Dumitru Erhan, Christian Szegedy, Scott Reed, Cheng-Yang Fu, and Alexander~C. Berg.
\newblock {SSD}: Single shot {MultiBox} detector.
\newblock In \emph{European Conference on Computer Vision}, pages 21--37, 2016.

\bibitem[{OpenAI}(2026)]{openai2026gpt61sol}
{OpenAI}.
\newblock {GPT-6.1 Sol}.
\newblock OpenAI model documentation, 2026.
\newblock Accessed: 2026-11-17.

\bibitem[Oquab et~al.(2023)Oquab, Darcet, Moutakanni, Vo, Szafraniec, Khalidov, Fernandez, Haziza, Massa, El-Nouby, et~al.]{oquab2023dinov2}
Maxime Oquab, Timoth{\'e}e Darcet, Th{\'e}o Moutakanni, Huy Vo, Marc Szafraniec, Vasil Khalidov, Pierre Fernandez, Daniel Haziza, Francisco Massa, Alaaeldin El-Nouby, et~al.
\newblock Dinov2: Learning robust visual features without supervision.
\newblock \emph{arXiv preprint arXiv:2304.07193}, 2023.

\bibitem[Qiu et~al.(2007)Qiu, Sugihara, and von~der Heydt]{qiu2007figureground}
Fangtu~T. Qiu, Tadashi Sugihara, and R{\"u}diger von~der Heydt.
\newblock Figure-ground mechanisms provide structure for selective attention.
\newblock \emph{Nature Neuroscience}, 10\penalty0 (11):\penalty0 1492--1499, 2007.

\bibitem[Radford et~al.(2021)Radford, Kim, Hallacy, Ramesh, Goh, Agarwal, Sastry, Askell, Mishkin, Clark, Krueger, and Sutskever]{radford2021clip}
Alec Radford, Jong~Wook Kim, Chris Hallacy, Aditya Ramesh, Gabriel Goh, Sandhini Agarwal, Girish Sastry, Amanda Askell, Pamela Mishkin, Jack Clark, Gretchen Krueger, and Ilya Sutskever.
\newblock Learning transferable visual models from natural language supervision.
\newblock In \emph{Proceedings of the 38th International Conference on Machine Learning}, pages 8748--8763, 2021.

\bibitem[Su et~al.(2024)Su, Ahmed, Lu, Pan, Bo, and Liu]{su2024roformer}
Jianlin Su, Murtadha Ahmed, Yu Lu, Shengfeng Pan, Wen Bo, and Yunfeng Liu.
\newblock Roformer: Enhanced transformer with rotary position embedding.
\newblock \emph{Neurocomputing}, 568:\penalty0 127063, 2024.

\bibitem[Sun et~al.(2025)Sun, Huang, Liu, Wu, Xu, Li, and Liu]{sun2025t2v}
Kaiyue Sun, Kaiyi Huang, Xian Liu, Yue Wu, Zihan Xu, Zhenguo Li, and Xihui Liu.
\newblock T2v-compbench: A comprehensive benchmark for compositional text-to-video generation.
\newblock In \emph{2025 IEEE/CVF Conference on Computer Vision and Pattern Recognition (CVPR)}, pages 8406--8416. IEEE, 2025.

\bibitem[Vandersanden et~al.(2026)Vandersanden, Gadelha, Huang, Jeong, and Gryaditskaya]{vandersanden2026emvid}
Jente Vandersanden, Matheus Gadelha, Chun-Hao~P. Huang, Hyeonho Jeong, and Yulia Gryaditskaya.
\newblock {EM-Vid}: Training-free entity-centric memory for efficient and consistent multi-shot video generation.
\newblock arXiv preprint arXiv:2605.23610, 2026.

\bibitem[Virtanen et~al.(2020)Virtanen, Gommers, Oliphant, Haberland, Reddy, Cournapeau, Burovski, Peterson, Weckesser, Bright, {van der Walt}, Brett, Wilson, Millman, Mayorov, Nelson, Jones, Kern, Larson, Carey, Polat, Feng, Moore, {VanderPlas}, Laxalde, Perktold, Cimrman, Henriksen, Quintero, Harris, Archibald, Ribeiro, Pedregosa, {van Mulbregt}, and {SciPy 1.0 Contributors}]{virtanen2020scipy}
Pauli Virtanen, Ralf Gommers, Travis~E. Oliphant, Matt Haberland, Tyler Reddy, David Cournapeau, Evgeni Burovski, Pearu Peterson, Warren Weckesser, Jonathan Bright, St{\'e}fan~J. {van der Walt}, Matthew Brett, Joshua Wilson, K.~Jarrod Millman, Nikolay Mayorov, Andrew R.~J. Nelson, Eric Jones, Robert Kern, Eric Larson, C~J Carey, {\.I}lhan Polat, Yu Feng, Eric~W. Moore, Jake {VanderPlas}, Denis Laxalde, Josef Perktold, Robert Cimrman, Ian Henriksen, E.~A. Quintero, Charles~R. Harris, Anne~M. Archibald, Ant{\^o}nio~H. Ribeiro, Fabian Pedregosa, Paul {van Mulbregt}, and {SciPy 1.0 Contributors}.
\newblock {{SciPy} 1.0: Fundamental Algorithms for Scientific Computing in Python}.
\newblock \emph{Nature Methods}, 17:\penalty0 261--272, 2020.

\bibitem[Wan et~al.(2025)Wan, Wang, Ai, Wen, Mao, Xie, Chen, Yu, Zhao, Yang, et~al.]{wan2025wan}
Team Wan, Ang Wang, Baole Ai, Bin Wen, Chaojie Mao, Chen-Wei Xie, Di Chen, Feiwu Yu, Haiming Zhao, Jianxiao Yang, et~al.
\newblock Wan: Open and advanced large-scale video generative models.
\newblock \emph{arXiv preprint arXiv:2503.20314}, 2025.

\bibitem[Wang et~al.(2024)Wang, He, Li, Li, Yu, Ma, Li, Chen, Chen, Wang, et~al.]{wang2024internvid}
Yi Wang, Yinan He, Yizhuo Li, Kunchang Li, Jiashuo Yu, Xin Ma, Xinhao Li, Guo Chen, Xinyuan Chen, Yaohui Wang, et~al.
\newblock Internvid: A large-scale video-text dataset for multimodal understanding and generation.
\newblock In \emph{International Conference on Learning Representations}, pages 42055--42079, 2024.

\bibitem[Wang et~al.(2026)Wang, Yao, Qiu, Shi, Wang, He, Hong, and Li]{weaveforcing2026}
Ziyi Wang, Junchi Yao, Heqian Qiu, Wenbo Shi, Chengjiu Wang, Jinyang He, Binkai Hong, and Hongliang Li.
\newblock {Weave Forcing}: Compositional memory routing for interactive long video generation.
\newblock arXiv preprint arXiv:2610.03510, 2026.

\bibitem[Wu et~al.(2026)Wu, Feng, Zhang, Qin, Li, Fan, Liu, An, Huang, Xu, and Yang]{wu2026echoforcing}
Mingqiang Wu, Weilun Feng, Zhefeng Zhang, Haotong Qin, Yuqi Li, Guoxin Fan, Xiaokun Liu, Zhulin An, Libo Huang, Yongjun Xu, and Chuanguang Yang.
\newblock {Echo-Forcing}: A scene memory framework for interactive long video generation.
\newblock arXiv preprint arXiv:2605.16003, 2026.

\bibitem[Xiao et~al.(2024)Xiao, Wu, Xu, Dai, Hu, Lu, Zeng, Liu, and Yuan]{xiao2024florence2}
Bin Xiao, Haiping Wu, Weijian Xu, Xiyang Dai, Houdong Hu, Yumao Lu, Michael Zeng, Ce Liu, and Lu Yuan.
\newblock {Florence-2}: Advancing a unified representation for a variety of vision tasks.
\newblock In \emph{Proceedings of the IEEE/CVF Conference on Computer Vision and Pattern Recognition}, pages 4818--4829, 2024.

\bibitem[Yang et~al.(2026)Yang, Huang, Chu, Xiao, Zhao, Wang, Li, Xie, Chen, Lu, Han, and Chen]{longlive2025}
Shuai Yang, Wei Huang, Ruihang Chu, Yicheng Xiao, Yuyang Zhao, Xianbang Wang, Muyang Li, Enze Xie, Yingcong Chen, Yao Lu, Song Han, and Yukang Chen.
\newblock {LongLive}: Real-time interactive long video generation.
\newblock In \emph{International Conference on Learning Representations}, 2026.

\bibitem[Yin et~al.(2024)Yin, Gharbi, Zhang, Shechtman, Durand, Freeman, and Park]{yin2024one}
Tianwei Yin, Micha{\"e}l Gharbi, Richard Zhang, Eli Shechtman, Fredo Durand, William~T Freeman, and Taesung Park.
\newblock One-step diffusion with distribution matching distillation.
\newblock In \emph{2024 IEEE/CVF Conference on Computer Vision and Pattern Recognition (CVPR)}, pages 6613--6623. IEEE, 2024.

\bibitem[Yin et~al.(2025)Yin, Zhang, Zhang, Freeman, Durand, Shechtman, and Huang]{causvid2025}
Tianwei Yin, Qiang Zhang, Richard Zhang, William~T. Freeman, Fredo Durand, Eli Shechtman, and Xun Huang.
\newblock From slow bidirectional to fast autoregressive video diffusion models.
\newblock In \emph{Proceedings of the IEEE/CVF Conference on Computer Vision and Pattern Recognition}, pages 22963--22974, 2025.

\bibitem[Zhu et~al.(2021)Zhu, Huang, Deng, Ye, Huang, Chen, Zhu, Yang, Lu, Du, and Zhou]{zhu2021webface260m}
Zheng Zhu, Guan Huang, Jiankang Deng, Yun Ye, Junjie Huang, Xinze Chen, Jiagang Zhu, Tian Yang, Jiwen Lu, Dalong Du, and Jie Zhou.
\newblock {WebFace260M}: A benchmark unveiling the power of million-scale deep face recognition.
\newblock In \emph{Proceedings of the IEEE/CVF Conference on Computer Vision and Pattern Recognition}, pages 10492--10502, 2021.

\end{thebibliography}
}

\appendix

\section{Memory Construction and Read Details}
\label{app:memory}

\subsection{Detailed Procedures}
\label{app:procedures}
Algorithm~\ref{alg:framework} merges memory formation and generation into one loop over chunks.
Algorithms~\ref{alg:formation} and~\ref{alg:generation} separate the two.
Memory formation runs after each chunk is completed and commits the scene's records when the scene ends.
Retrieval runs once before the scene, and the router takes each person's face from the scene that introduced the person and the head hair and garments from the requested version, omitting the components that the content prompt changes.
\begin{algorithm}[t]
\caption{\textbf{Memory formation for scene $s$.}}
\label{alg:formation}
\small
\begin{algorithmic}[1]
\Require prompt $p_s$, memory $\mathcal M_{s-1}$, quota $k_s$
\Ensure memory $\mathcal M_s$
\State $\mathcal T_s\gets\operatorname{LLM}(p_s)$ \Comment{detector queries}
\State $\mathcal I^{\mathrm{bg}}_s\gets\varnothing$
\For{each completed chunk $c$ of scene $s$}
  \State Keep clean KV and attention $a_{c,\cdot}$ \Comment{clean pass}
  \State Decode $c$ and take one frame per latent step $t$
  \State Detect with $\mathcal T_s$ and track to local IDs $j$
  \For{each accepted observation $(j,t)$}
    \State $\mathcal I^{\mathrm{ob}}_{s,j}\gets\mathcal I^{\mathrm{ob}}_{s,j}\cup\big(\{t\}\times\Omega_{s,j,t}\big)$ \Comment{Eq.~\eqref{eq:object_memory_indices}}
  \EndFor
  \State $\mathcal E_{s,c}\gets$ tokens of $c$ outside all boxes
  \State $\mathcal I^{\mathrm{bg}}_s\gets\operatorname{TopK}_{k_s}\big(\mathcal I^{\mathrm{bg}}_s\cup\mathcal E_{s,c};\widetilde a\big)$ \Comment{Eq.~\eqref{eq:background_memory_indices}}
\EndFor
\State Gather K/V at all kept indices from every layer
\State $\mathcal D_s\gets\operatorname{LLM}(\text{detection lists of scene } s)$ \Comment{descriptions}
\State $\mathcal M_s\gets\mathcal M_{s-1}\cup\operatorname{Records}\big(\mathcal I^{\mathrm{ob}}_{s,\cdot},\mathcal I^{\mathrm{bg}}_s,\mathcal D_s\big)$ \Comment{commit}
\end{algorithmic}
\end{algorithm}

\begin{algorithm}[t]
\caption{\textbf{Retrieval and generation for scene $s$.}}
\label{alg:generation}
\small
\begin{algorithmic}[1]
\Require prompt $p_s$, reuse query $q_s$, memory $\mathcal M_{s-1}$
\Ensure scene $X_s$
\State $\widehat{\mathcal A}_s\gets\operatorname{LLM}\big(p_s,q_s,\operatorname{Catalog}(\mathcal M_{s-1})\big)$ \Comment{Eq.~\eqref{eq:language_memory_routing}}
\State Check that each selected record has KV \Comment{validate}
\State $\mathcal I\gets$ union of the selected token sets \Comment{deduplicate}
\State $\mathcal K_s\gets$ original K/V at $\mathcal I$ from every layer
\State $p_s^{+}\gets\operatorname{Refine}(p_s,\widehat{\mathcal A}_s)$ \Comment{fixed template}
\State Clear the native KV cache and move $\mathcal K_s$ to GPU
\For{each chunk $c$ of scene $s$}
  \For{each denoising step}
    \State Attend to $[K^m_d;K^n_d]$ at each layer $d$ \Comment{Eq.~\eqref{eq:app_mix}}
  \EndFor
  \State Correct the latent amplitude \Comment{Eq.~\eqref{eq:app_amp}}
  \State Update memory with $c$ \Comment{Alg.~\ref{alg:formation}, l.~4--11}
\EndFor
\State Commit scene $s$ \Comment{Alg.~\ref{alg:formation}, l.~13--15}
\end{algorithmic}
\end{algorithm}

\subsection{Source Alignment and Object Regions}
\label{app:objects}
Our backbone is the Self Forcing DMD EMA model~\cite{huang2025selfforcing, yin2024one} based on Wan2.1-T2V-1.3B~\cite{wan2025wan}, with 30 transformer layers, 12 attention heads, and head dimension 128.
At $832\times480$ resolution, one latent frame corresponds to a $30\times52$ token grid, and each chunk contains three latent frames.
The causal VAE maps latent frames to RGB frames with an offset at the first frame.
After each chunk is completed, the observer decodes it and selects one representative RGB frame per latent frame using this mapping.
Sources are indexed across the session, so their indices remain unique across chunks and scenes.
A detected box covers every token cell that it overlaps.
Detections are associated only within a scene, so they do not establish identity across scenes.
The memory keeps valid detections from the entire scene.

\paragraph{Detectors.}
We run two detectors on every representative frame.
Florence-2-base~\cite{xiao2024florence2} performs caption-to-phrase grounding for each object-category query that the LLM extracts from the content prompt (people, faces, head hair, garments, and mentioned objects).
Its grounding output has no calibrated confidence. We also run SSDLite320 with a MobileNetV3-Large backbone~\cite{liu2016ssd,howard2019mobilenetv3} trained on COCO~\cite{lin2014coco}, which detects all 80 COCO categories with confidence scores.
SSDLite keeps detections with confidence of at least 0.25 after non-maximum suppression at an IoU of 0.55.
Each detection keeps its detector, label, and box, and the detections of both detectors enter tracking (Appendix~\ref{app:tracking}) and record construction (Appendix~\ref{app:llm_components}).

\paragraph{What a region contains.}
An object region is a set of latent timesteps and token positions, not a separate object-only feature.
A box can include some background or a neighboring object, and attention can mix information from outside the box into the stored KV.
Keeping the full region therefore means that every original token inside the region is retained. It does not mean that the region is semantically isolated.

\subsection{Scene-Local Object Tracking}
\label{app:tracking}
Tracking links detections within a scene so that all observations of one object contribute to the same memory record.
Its inputs are the detections aligned to latent frames (class, box, and confidence when available) and the current track state.
Its outputs are scene-local instance IDs and an observation state for each detection: valid, missing, or ambiguous.
These IDs are separate from the names that refer to a person across scenes.
New detections are associated with active tracks, and accepted matches keep their IDs.
Only accepted observations add object regions, and a missing or ambiguous observation never creates a box.
Accepted boxes are projected onto the token grid of their latent frame to obtain the object-memory indices.
We implement the tracker ourselves as a conservative geometric associator (Algorithm~\ref{alg:tracking}).
Each track predicts its next box at constant velocity, and a detection is matched to a same-class track only when the two are each other's unique candidate by box overlap.
Unresolved cases are marked ambiguous rather than forced into an identity.

\begin{algorithm}[t]
\caption{\textbf{Scene-local object tracking for one frame.}}
\label{alg:tracking}
\small
\begin{algorithmic}[1]
\Require detections $\mathcal D_t$ of frame $t$, recent tracks $\mathcal T$ of scene $s$
\State predict the box $\hat b_\tau$ of each $\tau\in\mathcal T$ at constant velocity
\State $C_d\gets$ same-class $\tau$ with $\operatorname{IoU}(d,\hat b_\tau)\ge0.1$, for each $d\in\mathcal D_t$
\State match $d$ to $\tau$ when each is the other's unique candidate
\State mark other $d$ near a same-class track ambiguous
\State start a new track for each remaining $d$
\State mark unmatched tracks missing
\end{algorithmic}
\end{algorithm}

\subsection{Attention Scores for Background Selection}
\label{app:bg_scores}
Background tokens are ranked by the attention they receive during the clean re-caching pass.
Our scorer samples layers 0, 14, and 29, all attention heads, and a $6\times10$ spatial grid of queries per latent frame, and it scores only keys from the completed chunk.
The softmax denominator still includes every key that the scoring pass can attend to, not only background candidates.
Object regions are excluded after scoring, using the boxes detected at each token's own latent frame.
A detected box excludes the overlapping background candidates even if its track association is unresolved.
Within each chunk, the remaining scores are divided by their mean, which keeps their order and makes chunks comparable before they are merged.
The background quota is shared across all chunks of a scene: each update merges the new candidates with the retained set and keeps the highest normalized scores, breaking ties by token identifier.
An empty candidate set adds nothing, and missing, negative, or non-finite scores are treated as failures rather than replaced by uniform scores.
Scores are fixed once observed and are not updated by later recall requests.
The same selected indices gather K/V at every transformer layer.

\subsection{Token Regions and Attention Visualization}
At $832\times480$, the token grid is $52\times30$, so one token covers $16\times16$ RGB pixels.
Box visualizations show every grid cell that overlaps the box, and the face-and-hair illustration expands the displayed box outward to this grid.
The attention heatmaps use the saved clean re-caching scores for keys of the displayed latent frame, drawn as one constant color per token.
Log scores are mapped between their within-frame 1st and 99th percentiles and clipped at these limits, which changes only the color scale.
The heatmap is drawn before object regions are excluded.
The overlays illustrate token selection and are not human-verified masks.

\subsection{Memory Integration and Scene Commit}
\label{app:integration}
The memory integrator connects stored KV to the records used for retrieval.
Its inputs are tracked object regions, selected background indices, the original layer-wise K/V, their source coordinates, and observed descriptions.
Its outputs are object and background records with catalog entries, including whether each record's KV is available.
A record is addressed by its source scene and scene-local ID, and each scene has a dedicated background ID.
During a scene, object evidence accumulates and the background selection is updated as chunks complete.
Original KV is gathered at the selected source and grid indices, and tokens with identical provenance are stored once.
When the scene ends, each record is paired with its scene, type, local ID, name, and observed appearance, and the records are committed without overwriting earlier scene-specific appearances.

At recall time, the integrator converts the selected records and the requested region operations into the KV supplied to generation.
It removes overlaps, merges compatible regions, and subtracts regions that the request excludes.
A containment relation, such as a face inside a person region, is used only when both regions belong to the same person and the same source version.
For example, selecting a person and their face from the same source observation keeps their shared tokens once.
Duplicates are identified by their original token provenance, not by equal image coordinates in different scenes or timesteps, and K/V values are never averaged or synthesized.
We also implement the integrator ourselves (Algorithm~\ref{alg:integration}).
It accepts only committed records of earlier scenes, reports a selection without stored KV instead of substituting another memory, and reads a token shared by several records once.

\begin{algorithm}[t]
\caption{\textbf{Record validation and KV integration.}}
\label{alg:integration}
\small
\begin{algorithmic}[1]
\Require selections $\widehat{\mathcal A}_s$, memory $\mathcal M_{s-1}$
\State report each selection that is not a committed earlier record with KV \Comment{no substitute}
\State $\mathcal I\gets$ union of the selected token keys (source frame, grid cell)
\State $\mathcal K_s\gets$ original K/V at $\mathcal I$ from every layer
\end{algorithmic}
\end{algorithm}

\subsection{Read-Time Mixing}
\label{app:mixing}
Stored memory records reside on CPU, and selected KV is transferred to GPU for generation.
Stored records keep their original K/V and positional encoding, and the selected records share the attention context with the ongoing scene's KV.
Let $d$ index a transformer layer ($\ell$ remains the role index of Section~\ref{sec:problem_formulation}).
For layer-$d$ queries $Q_d$, ongoing-scene KV $(K^n_d,V^n_d)$, and selected memory KV $(K^m_d,V^m_d)$, we compute
\begin{equation}
\begin{aligned}
O^n_d&=\mathrm{Attn}(Q_d,K^n_d,V^n_d),\\
O^m_d&=\mathrm{Attn}\big(Q_d,[K^m_d;K^n_d],[V^m_d;V^n_d]\big),\\
O_d&=O^n_d+\alpha_{s,d}\big(O^m_d-O^n_d\big),
\end{aligned}
\label{eq:app_mix}
\end{equation}
where $\alpha_{s,d}\in[0,1]$ sets the memory influence for scene $s$ at layer $d$.
With $\alpha_{s,d}=0$ only the ongoing-scene attention is computed, and with $\alpha_{s,d}=1$ only the memory-augmented attention is computed.
The coefficient never scales or edits the stored K/V.

\subsection{Latent and Decoder Settings}
\label{app:latent}
Two inference settings support generation without changing the generator weights.
First, when a latent frame's amplitude is too small, we rescale it.
Let $\mu$ and $\sigma$ be the mean and population standard deviation of a latent frame over channels and positions.
For a threshold $\tau$,
\begin{equation}
\hat z'=
\begin{cases}
\mu+(\hat z-\mu)\,\tau/\sigma, & 0<\sigma<\tau,\\
\hat z, & \sigma\ge\tau,
\end{cases}
\label{eq:app_amp}
\end{equation}
and a frame with zero variance is reported as a failure.
The corrected latent is used for both re-caching and RGB decoding.
Here $\sigma$ is the measured latent amplitude, not a change to the noise schedule.
Second, at scene boundaries the decoder clears its cached feature values while keeping its temporal and normalization bookkeeping. This is separate from clearing the attention KV or changing RoPE~\cite{su2024roformer} coordinates.

\subsection{LLM Components}
\label{app:llm_components}
Three steps use an LLM: the object-category queries for the detector, record construction at scene commit, and routing.
All three use \texttt{GPT-6.1 Sol high} and receive only text: the content prompt, the reuse query, the detection lists (the detector, label, and box of each detection), and the catalog.
They never see images.
Each response is a JSON object that a program validates before use: every referenced detection and record must exist, and a record can be selected only if its KV is available.
Prompt refinement uses a fixed template rather than an LLM.
For every evaluated history, we confirmed through an agreement check that rerunning record construction and routing on the recorded inputs reproduces the records and selections used for generation.
Appendix~\ref{app:llm_prompts} lists the instructions given to the LLM.

\subsection{Resource Accounting}
\label{app:resources}
Table~\ref{tab:app_resources} reports GPU memory and retained memory for each method on NVIDIA RTX A6000 GPUs.
Peak and reserved memory come from PyTorch's allocator statistics, so the peak is the allocated peak rather than the process peak reported by \texttt{nvidia-smi}.
Retained memory is the deduplicated sum of KV, memory banks, and buffers, excluding model weights, and the table reports its maximum over the history.
It is measured at scene boundaries rather than continuously, so it is not a continuous peak.
\begin{table}[t]
\centering
\footnotesize
\setlength{\tabcolsep}{5pt}
\caption{\textbf{GPU memory and retained KV.} Peak: PyTorch's peak allocated memory per GPU. Reserved: memory reserved by PyTorch's allocator. Retained: the deduplicated sum of KV, memory banks, and buffers at scene boundaries, excluding model weights. All values in GiB on NVIDIA RTX A6000 GPUs. Lowest in bold, second lowest underlined.}
\label{tab:app_resources}
\begin{tabular}{l ccc}
\toprule
Method & Peak & Reserved & Max retained \\
\midrule
Self-Forcing  & \textbf{15.4} & \underline{21.9} & \phantom{0}\textbf{5.7} \\
SlotMemory    & 30.5 & 46.6 & \phantom{0}\underline{6.1} \\
Echo-Forcing  & 35.6 & 45.3 & 24.0 \\
LongLive-RAG  & \underline{17.1} & \textbf{20.6} & 64.4 \\
Weave Forcing & 20.5 & 26.6 & 24.6 \\
Ours          & 31.8 & 32.0 & 28.6 \\
\bottomrule
\end{tabular}
\end{table}

\section{Scenario Design and Authoring}
\label{app:benchmark_design}

\subsection{Six-Scene Histories and Source Assignments}
Each scenario follows two people, P1 and P2, through six scenes and five places (L1--L5).
Scene~1 introduces P1 at L1, and Scene~2 introduces P2 at L2.
Scene~3 shows P1 at L3 with one requested change to clothing or hair, and Scene~4 shows P2 at L4 with one requested change.
Scene~5 brings both people together in their initial appearances at L5.
Scene~6 combines requested appearances with a past background, following the assignments in Table~\ref{tab:app_sources}.

\begin{table}[t]
\centering
\footnotesize
\setlength{\tabcolsep}{3pt}
\caption{\textbf{Requested sources for the Scene~6 branches.} Scenes~1--2 give the initial versions and Scenes~3--4 the modified versions.}
\label{tab:app_sources}
\begin{tabular}{cccc}
\toprule
Branch & P1 source & P2 source & Background source \\
\midrule
A & Scene 1 (initial) & Scene 2 (initial) & Scene 4 \\
B & Scene 1 (initial) & Scene 4 (modified) & Scene 4 \\
C & Scene 3 (modified) & Scene 2 (initial) & Scene 4 \\
D & Scene 3 (modified) & Scene 4 (modified) & Scene 4 \\
E & Scene 3 (modified) & Scene 4 (modified) & Scene 1 \\
\bottomrule
\end{tabular}
\end{table}

Branches A--D cover all four combinations of the two people's versions in the same background, and branches D and E differ only in the background source.
All branches share the same Scene~6 content prompt, and only the reuse query changes the requested sources.
Every branch is generated independently from Scene~1, so each of the 50 histories per method and query type has its own Scenes~1--5, and each branch is evaluated against the sources in its own history.

\subsection{Motifs, Attributes, and Query Wording}
We adapt human-centered roles, places, and simple activities from MemFlow~\cite{memflow2025} (repository KlingAIResearch/MemFlow, commit \texttt{7ed5147}, file \texttt{prompts/interactive\_benchmark.jsonl}) and write new connected requests. We do not reuse MemFlow's evaluation set.
Each content prompt has 30--36 words and at most 40, and the reuse query is a separate field.
Scenes use simple visible activities, and faces and edited regions are required to be visible.
Requested changes cover clothing color, garment type, layers, sleeves, and hair, whereas glasses, facial hair, age, and body shape are not changed.
Explicit queries name a person and a scene number.
Natural queries refer to the same past observation through its place or activity, without repeating the requested appearance.
Content prompts for recall scenes do not restate the clothing, hair, or background layout to be recalled.
A separate evaluation specification, not shown to any method, records the intended roles, sources, changes, preserved attributes, and exclusions for every scene.

\subsection{Authoring and Text Checks}
One authoring request produces one scenario with its six scenes and five Scene~6 branches.
The authoring prompt contains the motif catalog, twenty assignment cards, query rules, and the output schema, and it is reproduced in Appendix~\ref{app:authoring_prompt}.
Automatic text checks cover word counts, source assignments, the equivalence of explicit and natural queries, the shared Scene~6 content, and the visibility of requested changes.
We keep the final text of every scenario together with its hash.

\section{Evaluation Details}

\begin{figure*}[t]
\centering
\includegraphics[width=\textwidth]{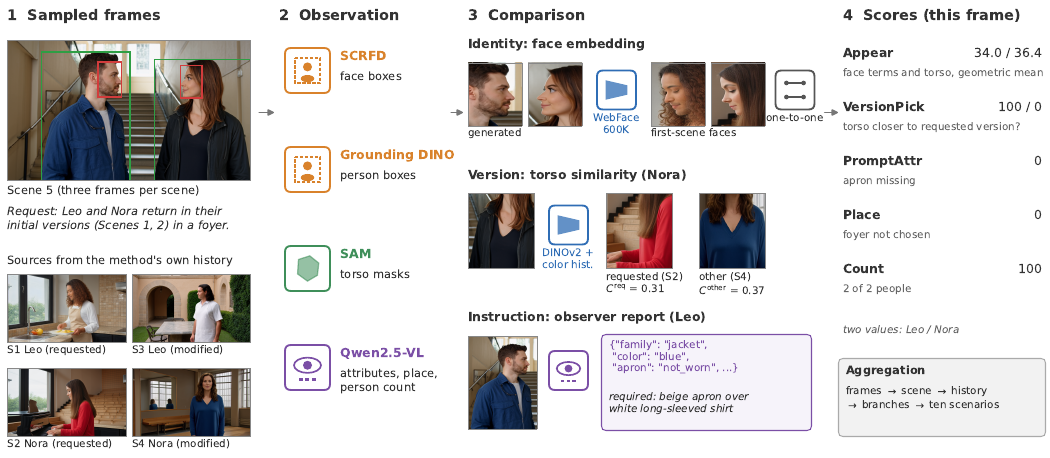}
\caption{\textbf{Evaluation pipeline on one generated frame.} The example is a Scene~5 frame of a Weave Forcing history (explicit queries), in which Leo and Nora should return in their initial versions. Detectors and the segmenter localize faces, people, and torsos, and the VLM observer reports visible attributes, the place, and the person count. Faces are compared with each person's first-scene face and matched one-to-one, torsos are compared with both versions of the person in the method's own history, and the observer report is scored against the written requirements. The values are the actual scores of this frame. Model icons are schematic, and the modified-version thumbnails are frames from Scenes~3 and~4 of the same history.}
\label{fig:eval_pipeline}
\end{figure*}

Figure~\ref{fig:eval_pipeline} follows one generated frame through the evaluation.

\subsection{Observer and Models}
\paragraph{Frames.}
From each scene we take three frames, near its beginning, middle, and end (about 10\%, 50\%, and 90\% of its length).
Face, torso, background, and person-count measurements use all three frames, and attribute, prohibited-item, place, and activity observations use the middle frame.
Generated faces are matched to the requested people by maximizing the total face similarity under a one-to-one constraint, so each face represents at most one person and clothing never decides the match.

\paragraph{Models.}
\label{app:observer_v2}
SCRFD-10GF (\texttt{det\_10g.onnx}) detects faces and five landmarks with a $640\times640$ input and detection threshold 0.5~\cite{guo2022scrfd}.
The WebFace600K ResNet50 model (\texttt{w600k\_r50.onnx}) computes 512-dimensional embeddings from landmark-aligned $112\times112$ crops~\cite{deng2019arcface,zhu2021webface260m}.
Faces used to build a reference must also have detector confidence of at least 0.65.
Grounding DINO Swin-T provides person boxes with box and text thresholds of 0.25 and same-label NMS at IoU 0.5~\cite{liu2023groundingdino}.
SAM ViT-B provides masks~\cite{kirillov2023segmentanything}, and DINOv2 ViT-B/14 provides torso and background patch features~\cite{oquab2023dinov2}.

The visual observer is Qwen2.5-VL-7B~\cite{bai2025qwen25vl}, run locally in BF16 with FlashAttention-2~\cite{dao2024flashattention} on two RTX 3090 GPUs, with greedy decoding.
Each request contains one full frame or one person crop and a fixed neutral question (Appendix~\ref{app:observer_prompts}).
The observer never receives the method name, character names, expected values, the requested source, or the exclusions.
A response that fails to parse, is truncated, or contains a value outside the allowed options is not accepted as an observation.
We store raw responses, parsed fields, and image hashes.

\subsection{Identity-Aware Appearance}
\label{app:appearance_v2}

\paragraph{Face references.}
For each person, we build the face reference from the first scene in which that person appears.
Qualified face embeddings from the last two seconds of that scene are normalized, averaged, and normalized again, after removing faces that belong to depictions such as portraits.
A reference needs at least two qualified faces. Otherwise, a single qualified face from the same scene is used, and if none exists the reference is unknown.

\paragraph{Matching faces to roles.}
Let $e^0_\ell$ be the reference of role $\ell$ and $e^T_\nu$ a generated face indexed by $\nu$.
We find the one-to-one assignment $\pi$ that maximizes $\sum_\ell \cos(e^0_\ell,e^T_{\pi(\ell)})$ with SciPy's linear-sum assignment~\cite{virtanen2020scipy}.
Clothing and attributes never influence this assignment.
To link a face to a body, we enumerate all one-to-one face--body matchings consistent with the detected boxes and keep only links shared by every feasible matching, and other links remain unknown.

\paragraph{Facial terms and constants.}
Let $g_\ell=\cos(e^0_\ell,e^T_{\pi(\ell)})$ be the similarity between the generated face and the intended person's reference, $g^{\mathrm{other}}_\ell=\cos(e^0_{3-\ell},e^T_{\pi(\ell)})$ the similarity to the other person's reference, and $g_{12}=\cos(e^0_1,e^0_2)$ the similarity between the two references.
The facial terms of Eq.~\eqref{eq:appearance_index_v2} are
\begin{equation}
\begin{aligned}
F_\ell^{\mathrm{abs}}&=\sigma\left((g_\ell-c_0)/\tau_f\right),\\
F_\ell^{\mathrm{rel}}&=\sigma\left((g_\ell-g_\ell^{\mathrm{other}})/\tau_f\right),\\
F^{\mathrm{sep}}&=\sigma\left((c_0-g_{12})/\tau_f\right),
\end{aligned}
\label{eq:appearance_facial_terms_main}
\end{equation}
where $\sigma$ is the sigmoid function with its input clipped to $[-40,40]$.
The constants come from medians measured on control pairs before evaluation:
\begin{equation}
\begin{aligned}
\mu^+&=0.9364,\quad \mu^-=0.1200,\\
c_0&=(\mu^++\mu^-)/2=0.5282,\\
\tau_f&=(\mu^+-\mu^-)/(2\ln 9)=0.1858,
\end{aligned}
\label{eq:app_constants}
\end{equation}
where $\mu^+$ is the median similarity of 14{,}148 same-person pairs across frames and $\mu^-$ that of 36 pairs of different initial people.
With these values, a similarity equal to $\mu^+$ gives $F^{\mathrm{abs}}=0.9$ and a similarity equal to $\mu^-$ gives $0.1$.
FaceKeep in Scenes~3--4 is $100\,F^{\mathrm{abs}}_\ell$, averaged over the person who changes.

\paragraph{Torso similarity.}
Let $\phi^S_\ell$ and $\phi^T_\ell$ be the normalized torso descriptors of the requested source and the matched generated person, and $p^S_{\ell,b}$ and $p^T_{\ell,b}$ their HSV histograms with unit sum over bins $b$.
Then
\begin{equation}
C_\ell=\tfrac12\Big[\cos(\phi^S_\ell,\phi^T_\ell)+\textstyle\sum_b\sqrt{p^S_{\ell,b}\,p^T_{\ell,b}}\Big],
\label{eq:app_torso}
\end{equation}
which enters Eq.~\eqref{eq:appearance_index_v2} after clipping to $[0,1]$.
To obtain the torso, we take the person box and its SAM mask and remove everything above $\lceil y_{\mathrm{face,bottom}}+0.25\,h_{\mathrm{face}}\rceil$, where $y_{\mathrm{face,bottom}}$ and $h_{\mathrm{face}}$ are the bottom and height of the face box.
Removed pixels are filled with gray (RGB 127), and the crop is resized to $224\times224$ (bicubic for images, nearest for masks) and normalized with ImageNet statistics.
We average the normalized DINOv2 patch features whose foreground occupancy is at least 0.7 and normalize the mean. The CLS token is not used.
A torso with fewer than four such patches or fewer than 32 foreground pixels is unknown.
The HSV histogram uses $16\times4\times4$ bins over the masked torso.
When SAM returns an insufficient mask, we prompt it again with the person box and points derived from the face box.

\subsection{Requested Attributes and Versions}
\label{app:attributes}
\paragraph{PromptAttr.}
For role $\ell$ and registered attribute $\kappa$, let $y^\star_{\ell,\kappa}$ be the required value and $\hat y_{\ell,\kappa}$ the observed value.
A sufficiently visible attribute scores $100\cdot\mathbf{1}[\hat y_{\ell,\kappa}=y^\star_{\ell,\kappa}]$, and an attribute that is not visible stays unknown.
We average the known attributes of each role and then average the roles with at least one known attribute, so a role with more attributes does not receive more weight.
The scored attributes are garment type, color, sleeves, hair presence, hair length, hair texture, and whether an apron, blazer, vest, cardigan, or jacket is worn.
Garment type and color describe the outermost visible upper-body garment, so an apron, vest, cardigan, or jacket takes priority over a shirt underneath.
Color names are merged into basic colors (for example, navy and light blue are blue), and garment names are merged into types (for example, blouse, polo, and T-shirt are shirts, and a blazer is a jacket).
Clothing attributes need sufficient torso visibility, hair presence and length need a full view of the head, and texture can use a partial view.

\paragraph{VersionAttr.}
Let $D^{\mathrm{init}}_\ell$ and $D^{\mathrm{mod}}_\ell$ be the attributes with registered values in the initial and modified versions of person $\ell$.
VersionAttr uses only the attributes whose values differ,
\begin{equation}
D^{\Delta}_\ell=\{\kappa\in D^{\mathrm{init}}_\ell\cap D^{\mathrm{mod}}_\ell:\ y^{\mathrm{init}}_{\ell,\kappa}\neq y^{\mathrm{mod}}_{\ell,\kappa}\},
\label{eq:app_version_set}
\end{equation}
and compares the output with the value of the version that the current scene requires: the modified value in Scenes~3--4, the initial value in Scene~5, and the requested version in Scene~6.
Scoring and averaging follow PromptAttr.
A role whose versions differ in no registered attribute is N/A.

\paragraph{VersionPick.}
VersionPick (Eq.~\eqref{eq:version_pick}) reuses the torso descriptors of Appear.
For each matched role and frame in Scenes~5--6, $C^{\mathrm{req}}_\ell$ uses the torso of the requested version and $C^{\mathrm{other}}_\ell$ the torso of the other version, both taken from the method's own history.
It applies only to roles whose two versions differ in a registered clothing attribute, and it is unknown when either similarity is unknown.

\subsection{Prohibited Items, Person Count, and Missing Observations}
\label{app:missingness_v2}
\paragraph{Forbidden.}
Prohibited items come from the explicit ``no'' and ``without'' clauses of the current scene and from garment or headgear exclusions registered for each person.
A visible prohibited item scores zero, and its absence scores 100 only when the relevant region is sufficiently visible. Occlusion, an off-frame region, or an unmatched person leaves the item unknown.
Scene-level and person-level items are pooled into one list, and duplicates are removed.

\paragraph{Count.}
For an observed count $n$ and a requested count $n^\star$, Count is $100\cdot\min(n,n^\star)/\max(n,n^\star)$ and exact count is $100\cdot\mathbf{1}[n=n^\star]$.
For example, one person when two are requested scores 50, and three people score 66.7.
Portraits, statues, and mirror reflections are not counted as additional people.
The observer answers three neutral count questions for each frame: a short count question, a count-only question that first lists each person's position, and the count field of the full-frame observation (Appendix~\ref{app:observer_prompts}).
A count is accepted only when all available answers agree. Otherwise, Count is unknown for that frame.

\paragraph{Confirmed absence.}
A requested person is confirmed absent when at least two count answers report fewer people than requested and every visible face is already matched to a requested person.
A confirmed absence scores zero in Appear and in that person's attribute scores, while a person who simply could not be measured stays unknown.

\paragraph{Bounds.}
Let $R$ be the number of requested people, $O$ the set of measured roles, and $n_{\mathrm{unk}}$ the number of roles that remain unknown.
A frame's Appear range is
\begin{equation}
\begin{aligned}
\mathrm{Appear}^{\mathrm{lo}}&=\frac{\sum_{\ell\in O}\mathrm{Appear}_\ell}{R},\\
\mathrm{Appear}^{\mathrm{hi}}&=\frac{\sum_{\ell\in O}\mathrm{Appear}_\ell+100\,n_{\mathrm{unk}}}{R},
\end{aligned}
\label{eq:app_bounds}
\end{equation}
where a confirmed absent person contributes zero to both sums.
The frame value equals both bounds when $n_{\mathrm{unk}}=0$ and is unknown otherwise.
For example, one person at 70 and one confirmed absent person give 35. If the second person is instead unknown, the frame is unknown with range $[35,85]$.
Attribute and prohibited-item ranges are obtained in the same way by setting unknown items to zero or 100.
These ranges show the possible effect of unknown values and are not confidence intervals.

\subsection{Place and Activity}
\label{app:place}
Each scenario registers five places and the activities of its scenes.
For the middle frame of each scene, the observer chooses one place from all five place descriptions, ``none of these'', or ``cannot tell'', and one activity from the scenario's activities with the same two extra options.
The options are shuffled once per scenario and named by neutral codes, so the requested answer is never marked.
Choosing the requested option scores 100, choosing another option or ``none of these'' scores zero, and ``cannot tell'' is unknown.
Scenes~5 and~6 show the same two-person interaction, so either of their activity options is accepted in those scenes.
As a cross-check, CLIP ViT-L/14~\cite{radford2021clip} ranks the five place descriptions for the same frame, and the top-ranked place is scored in the same way.

\subsection{Background Correspondence}
\label{app:bgg}
BG-G compares the background of a Scene~6 frame with that of the requested source scene.
The excluded region is the union of person masks, face boxes, and accepted prop masks and boxes, dilated by $\lceil\min(H,W)/160\rceil$ pixels for an image of size $H\times W$ and filled with gray.
The image is resized without upsampling so that its long side is at most 518 pixels and both sides are multiples of 14, and patches that touch the excluded region or the image border are removed.
Normalized DINOv2 patches are matched as mutual nearest neighbors with cosine similarity of at least 0.65.
Among these $N_{\mathrm{pair}}$ matches, $N_{\mathrm{in}}$ pass a homography estimated with OpenCV USAC-MAGSAC~\cite{barath2020magsacplusplus} (24-pixel threshold, 1{,}000 iterations, confidence 0.999) and a forward/backward reprojection check of 24 pixels.
As a chance reference, we repeat the estimation for 63 shuffled target coordinates and take the 95th-percentile inlier count $I^{\mathrm{null}}_{.95}$ and the permutation $p$-value $p_{\mathrm{perm}}=(1+\#\{\text{null}\ge N_{\mathrm{in}}\})/64$.
The layout term is
\begin{equation}
L_{\mathrm{geom}}=\left[\frac{N_{\mathrm{in}}-I^{\mathrm{null}}_{.95}}{N_{\mathrm{pair}}-I^{\mathrm{null}}_{.95}}\right]_0^1
\label{eq:app_bgg}
\end{equation}
when the denominator is positive, $p_{\mathrm{perm}}\le0.05$, and $N_{\mathrm{in}}\ge8$, and zero otherwise.
BG-G is 100 times the mean cosine similarity of the inliers times $L_{\mathrm{geom}}$, and it also requires the inliers' convex hull to cover at least 3\% of each image.
A frame with fewer than eight valid patches or less than 3\% possible coverage is unknown, while an observable frame with poor alignment scores zero.
Random draws use fixed seeds.

\subsection{Aggregation and Video Quality}
\label{app:aggregation}
Each metric is first reduced within a scene (the mean of its measurable frames, or the middle frame for observer-based metrics) and then within a history, where every applicable scene has equal weight.
We then average repeated generations of the same branch, average the five branches of a scenario, and give each of the ten scenarios equal weight. Explicit and natural queries are reported separately.
Missing histories are not imputed from other branches.
VBench~\cite{huang2023vbench} evaluates each scene clip with its original content prompt. Table~\ref{tab:main_v2} reports aesthetic quality (CLIP ViT-L/14~\cite{radford2021clip} with a linear predictor), imaging quality (MUSIQ~\cite{ke2021musiq}), and overall consistency, the text--video alignment measured with ViCLIP~\cite{wang2024internvid}.

\section{Full Results}
\label{app:full_results_v2}
Table~\ref{tab:app_supplementary} reports the metrics that are not in the main table, together with VersionPick and Place for individual scenes, and Table~\ref{tab:app_coverage} reports how often each metric could be measured.
Each method has 50 independently generated histories per query type (ten scenarios with five branches each).
\begin{table*}[t]
\centering
\fontsize{9.0}{10}\selectfont
\setlength{\tabcolsep}{4pt}
\caption{\textbf{Supplementary results.} Explicit / Natural queries on a 0--100 scale (higher is better). FaceKeep is face similarity to the first-scene reference during Scenes~3--4. VersionPick and Place are also shown for single scenes. Best in bold, second best underlined.}
\label{tab:app_supplementary}
\begin{tabular}{l ccccccc}
\toprule
Method & Forbidden $\uparrow$ & Exact count $\uparrow$ & Activity $\uparrow$ & FaceKeep (S3--4) $\uparrow$ & VersionPick (S5) $\uparrow$ & VersionPick (S6) $\uparrow$ & Place (S6) $\uparrow$ \\
\midrule
Self-Forcing & 12.5 / 0.0 & 66.7 / 66.7 & 19.2 / 18.6 & \underline{62.3} / \underline{59.2} & 0.0 / 0.0 & \underline{60.0} / \underline{59.2} & 12.5 / 12.5 \\
Echo-Forcing & 29.2 / \underline{24.6} & \underline{98.3} / 93.6 & 64.3 / 55.7 & 16.9 / 17.2 & 24.3 / 41.7 & 49.3 / 50.3 & 8.9 / 18.0 \\
LongLive-RAG & 23.8 / 13.3 & 80.0 / 81.7 & 32.7 / 34.3 & \textbf{68.0} / \textbf{66.8} & 20.0 / 5.0 & 57.1 / 58.0 & 20.0 / 24.0 \\
SlotMemory & 15.8 / 15.0 & 83.3 / 81.3 & 57.0 / 43.7 & 33.9 / 37.4 & \underline{60.3} / 30.0 & 51.0 / 53.8 & 26.5 / 30.0 \\
Weave Forcing & \underline{30.0} / 21.4 & 92.9 / \underline{93.9} & \underline{74.5} / \underline{67.4} & 18.2 / 26.3 & 37.3 / \underline{50.2} & 58.5 / 50.2 & \underline{52.0} / \underline{52.0} \\
Ours & \textbf{87.5} / \textbf{87.5} & \textbf{98.7} / \textbf{98.7} & \textbf{80.9} / \textbf{80.7} & 35.9 / 37.5 & \textbf{93.7} / \textbf{91.0} & \textbf{94.7} / \textbf{95.0} & \textbf{78.0} / \textbf{87.5} \\
\bottomrule
\end{tabular}
\end{table*}

\begin{table}[t]
\centering
\small
\setlength{\tabcolsep}{4pt}
\caption{\textbf{Measurement coverage} over all methods and query types. Measured: at least one known value. Complete: every applicable value known. Range: mean gap between the bounds obtained by setting unknown values to 100 or zero.}
\label{tab:app_coverage}
\begin{tabular}{l cccc}
\toprule
Metric & Applicable & Measured & Complete & Range \\
\midrule
Appear & 608 & 605 & 481 & 3.1 \\
PromptAttr & 608 & 608 & 499 & 4.0 \\
VersionAttr & 608 & 608 & 542 & 4.0 \\
Forbidden & 499 & 482 & 275 & 41.8 \\
Count & 608 & 608 & 586 & 0.2 \\
BG-G & 608 & 608 & 608 & 0.0 \\
FaceKeep (S3--4) & 608 & 608 & 608 & 0.0 \\
VersionPick & 608 & 598 & 598 & 0.0 \\
Place & 608 & 608 & 608 & 0.0 \\
Activity & 608 & 608 & 608 & 0.0 \\
\bottomrule
\end{tabular}
\end{table}

\section{Qualitative Analysis}
\label{app:qualitative}
Figure~\ref{fig:motivation} shows a natural-query history of the scenario with Amir and Sofia (branch~C).
Each baseline fails there in a characteristic way, and Table~\ref{tab:app_diagnostics} shows that the same patterns hold across all evaluated histories.
Numbers below refer to explicit queries over the ten scenarios unless stated otherwise.
\begin{table}[t]
\centering
\footnotesize
\setlength{\tabcolsep}{2.5pt}
\caption{\textbf{Diagnostics behind the qualitative patterns.} Places: distinct places reported over Scenes~1--5 (five requested). One place: histories that show one place throughout. Distinct: separability of the two first-scene faces. The other columns use explicit queries. Best in bold, second best underlined (lower is better for One place).}
\label{tab:app_diagnostics}
\begin{tabular}{l ccccc}
\toprule
Method & Places & One place & Distinct & FaceKeep & Exact count \\
\midrule
Self-Forcing & 1.36 & 64\% & 10.3 & \underline{62.3} & 66.7 \\
Echo-Forcing & 2.60 & \textbf{0\%} & 15.2 & 16.9 & \underline{98.3} \\
LongLive-RAG & 1.08 & 92\% & 26.4 & \textbf{68.0} & 80.0 \\
SlotMemory & 1.45 & \underline{55\%} & 54.5 & 33.9 & 83.3 \\
Weave Forcing & \underline{3.42} & \textbf{0\%} & \underline{89.0} & 18.2 & 92.9 \\
Ours & \textbf{4.67} & \textbf{0\%} & \textbf{91.7} & 35.9 & \textbf{98.7} \\
\bottomrule
\end{tabular}
\end{table}

\paragraph{Self-Forcing repeats one scene.}
In Figure~\ref{fig:motivation}, Self-Forcing shows Amir in the cafe in every scene, including Scene~2, which should introduce Sofia in a garden, and the frames degrade as the history grows.
Across histories it shows 1.4 of the five requested places on average, and 64\% of its histories never leave the first place.
Its first two people are also hard to tell apart (10.3).
Repetition explains its scores: the face and the room stay the same, so FaceKeep (62.3) and BG-G (64.9) are high, while VersionAttr (10.2) and Place (20.2) are low, and its Scene~5 output is always closer to the most recent torso than to the requested initial one (VersionPick 0.0).

\paragraph{LongLive-RAG stays in one place.}
In the figure, Amir remains in the cafe through Scenes~1--4, Sofia never appears, and the orange shirt is never shown. Scenes~5--6 add people inside the same cafe.
In 92\% of its histories all five scenes show one place, and its Place score stays at chance (18.3).
Its FaceKeep of 68.0 therefore reflects an unchanged person rather than a successful edit, as its VersionAttr of 18.2 shows.

\paragraph{SlotMemory copies the background and carries items over.}
In the figure, every scene stays in the cafe, the apron remains in Scene~3 although the request asks for an orange shirt, and Scenes~5--6 contain an extra person and objects from earlier scenes.
More than half of its histories (55\%) show one place, which gives it the second-highest BG-G (59.3), and its exact count is 83.3.

\paragraph{Echo-Forcing shows the right number of people, but not the right people.}
In the figure, Scenes~5--6 contain two people as requested, but neither looks like Amir or Sofia.
Its exact count is 98.3, while its Appear is among the lowest (21.9), its first two people are hardly distinct (15.2), and its FaceKeep is 16.9.

\paragraph{Weave Forcing follows the text but loses identity.}
In the figure, Weave Forcing introduces Sofia correctly and reaches the requested places more often, but Scene~3 shows an apron instead of an orange shirt, and the later scenes replace the two people and add a third.
Its Place score is the second highest (65.5) and its first two people are distinct (89.0), but its FaceKeep is 18.2 and its VersionPick stays near chance (47.9).

\paragraph{Our method.}
In the figure, every scene shows the requested people, versions, and places: Amir in the orange shirt in the bookshop, both initial versions in the square, and Amir's modified version with Sofia's initial version in the living room.
Across histories it shows 4.7 of the five places and keeps the two people distinct (91.7).

Each pattern lowers a different column of Table~\ref{tab:main_v2}, which is why no single measurement suffices to judge selective recall.

\section{Selective Recall in a Simpler Setting}
\label{app:simple_setting}
\begin{figure*}[t]
\centering
\includegraphics[width=\textwidth]{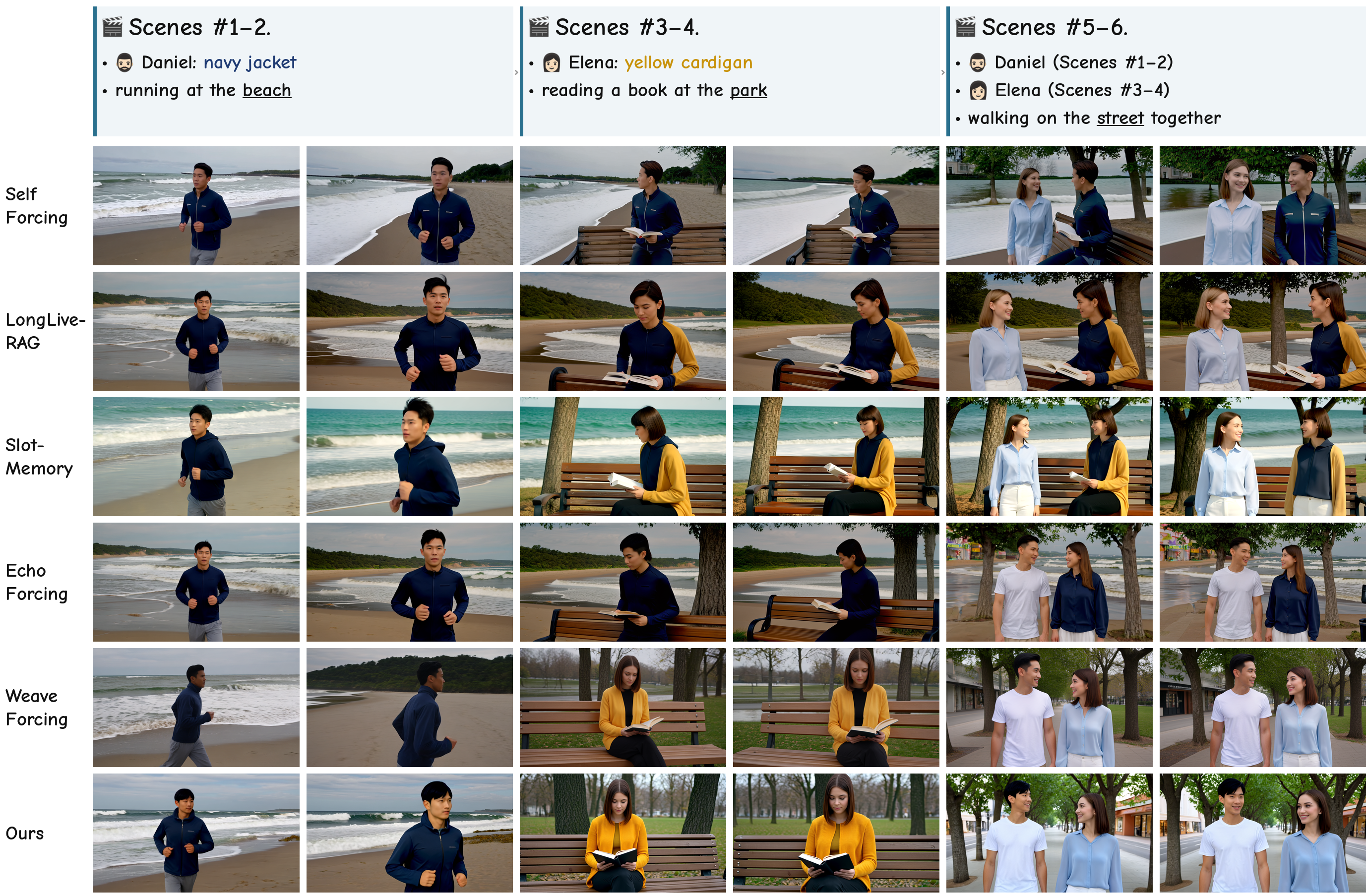}
\caption{\textbf{Generation in a simpler 60-second setting.} Each place continues for two scenes, and recall needs only the faces of earlier people. In this relatively continuous and simple setting, most methods produce clean frames and follow the requests more closely than in our scenarios, which require precise time-specific recall of individual elements. Each column shows one frame per scene.}
\label{fig:simple_setting}
\end{figure*}

To show how demanding our scenarios are, we also generate 60-second videos in a simpler six-scene setting (Figure~\ref{fig:simple_setting}).
Daniel jogs on a beach in Scenes~1--2, Elena reads on a park bench in Scenes~3--4, and the two walk along a tree-lined street in Scenes~5--6.
Each place continues for two consecutive scenes, and recall needs only the faces of the two people, so no earlier appearance has to be chosen.
When the setting is continuous and the elements to retrieve are simple, most methods generate videos that match the requests relatively well and with higher visual quality.
In this setting, every method keeps the frames clean and temporally consistent, without the progressive degradation seen in Figure~\ref{fig:motivation}, and Weave Forcing and our method follow all six scenes.
The other baselines mostly keep the beach as the background in later scenes, and some blend the clothing of the two people or show another person instead of Daniel in Scenes~5--6.
This contrasts with our scenarios, which require precise time-specific and element-wise recall: each person changes appearance and later returns to a specific version, and many baselines fall short there (Table~\ref{tab:main_v2}).

\section{Coverage and Robustness}
\label{app:robustness_v2}
\paragraph{Coverage.}
In Scene~6, 92 of 3{,}252 person-frame appearance measurements (2.8\%) could not be made, and every background and person-count frame was measured.

\paragraph{Frame substitution.}
An appearance measurement can fail in one sampled frame, for example when a face is turned away.
To test whether such gaps affect the results, we replace each unmeasurable frame with the nearest measurable sampled frame from the same part of the scene, using at most two candidates and never replacing a confirmed absence. Substitutes are chosen without looking at their scores.
This fills 54 of 84 unmeasurable Scene~6 frames and 20 of 33 unmeasurable Scene~5 cases.
Appear changes by at most 0.42 points, and our method remains first under both query types. The results in the main text use the original frames.

\paragraph{Independent place check.}
The CLIP-based place ranking (Appendix~\ref{app:place}) also places our method first and Weave Forcing second under both query types.
\section{Ablation Study}
\label{app:ablation_v2}
Each variant replaces one component of the full method with a simple alternative. Variants are not combined.
\begin{itemize}
\item \textbf{w.o.\ Prompt refinement} uses the original content instruction without the appearance descriptions added by refinement.
\item \textbf{w.o.\ Object-level memory extraction} supplies the whole-scene KV of the selected source scenes instead of the extracted object regions.
\item \textbf{w.o.\ LLM router} selects sources with BM25 over the same catalog, and the request parsing and refinement steps are unchanged.
\end{itemize}
Whole-scene memory supplies more KV than object regions, so this comparison is not matched in KV budget.

\paragraph{Six-scene study.}
The study uses one Scene~6 branch each (branches B--E), with both explicit and natural queries.
All variants share the same Scenes~1--2 and generate Scenes~3--6 themselves, so each variant builds its own memory of the modified versions.
Scenes are about 10 seconds long, same as the main evaluation, and the three frames of each scene are taken near its beginning, middle, and end.
Every metric is computed as in Section~\ref{sec:experiments} and Appendix~\ref{app:observer_v2}, and Table~\ref{tab:ablation_v2} pools the two query types.

\section{Limitations}
\label{app:limitations}
\paragraph{Facial similarity during attribute modification.}
In Scenes~3--4, our FaceKeep score is lower than that of Self-Forcing and LongLive-RAG (Table~\ref{tab:app_supplementary}).
These two baselines largely keep the earlier appearance unchanged, as their low VersionAttr shows, whereas our method applies the requested changes. Still, keeping the face fixed while clothing or hair changes remains difficult for our method.

\paragraph{Exact background layout.}
Our BG-G is lower than that of methods that keep one room throughout the history, although our Place scores are the highest. Our method returns to the requested type of place more often but reproduces its exact layout less closely.

\paragraph{What the measurements show.}
All metrics are computed from generated frames, so they show whether the output matches the requested source but not which historical KV the generator read.
The indices are continuous scores rather than calibrated identity probabilities, and the attribute vocabulary is coarse.

\paragraph{Dependence on perception models.}
The framework relies on an LLM router and an open-vocabulary detector, and the evaluation relies on face detection, segmentation, and a VLM observer, and errors in these models can affect individual memories or observations.
Prohibited items are often too occluded to confirm their absence, which widens the range of Forbidden.

\section{Visual Observer Prompts}
\label{app:observer_prompts}
The system and user instructions below are given verbatim. The observer never receives expected attributes, requested sources, character names, or method names.

\subsection{Person-Crop Observation}
Output limit: 520 tokens.
\begin{lstlisting}
System: You are a careful visual observer. Use only visible evidence. Return JSON only. Unknown is allowed. Do not infer appearance from research hypotheses or intended instructions.

User: Observe the main actual person in this image crop. Do not identify the person or infer intended appearance. Describe only directly visible physical facts. If another person overlaps so the intended main person's feature cannot be isolated, use unknown. Return JSON only with exactly these keys:
{"head_visibility":"full|partial|unknown","torso_visibility":"sufficient|partial|unknown","hands_visibility":"both|one|none|unknown","family":"apron|shirt|dress|sweater|hoodie|cardigan|jacket|vest|jersey|unknown","color":"red|orange|yellow|green|blue|purple|pink|brown|black|white|gray|beige|multicolor|unknown","sleeves":"long|short|none|unknown","hair_present":"present|absent|unknown","hair_length":"bald|short|medium|long|unknown","hair_texture":"straight|wavy|curly|coily|unknown","apron":"worn|not_worn|unknown","blazer":"worn|not_worn|unknown","vest":"worn|not_worn|unknown","cardigan":"worn|not_worn|unknown","jacket":"worn|not_worn|unknown","headgear":"hat|wig|other|none_visible|unknown","evidence":"brief visible evidence"}.
family and color refer to the principal visible outermost upper-body garment: a worn protective apron, waistcoat/vest, open cardigan, or jacket takes priority over an underlying shirt. Navy/light blue are blue; blouse/polo/T-shirt/button-up are shirt; tailored blazer is jacket. Sleeve length refers to the visible clothing covering the person's arms. A scarf or held object is not an outer garment. Distinguish absent hair from hair hidden by angle/hat/crop. Do not infer no apron or no hat just because the relevant torso or scalp is cut off: use unknown. Hair texture and length need visible evidence.
\end{lstlisting}

\subsection{Full-Frame Observation}
Output limit: 1150 tokens.
\begin{lstlisting}
System: You are a careful visual observer. Use only visible evidence. Return JSON only. Unknown is allowed. Do not infer appearance from research hypotheses or intended instructions.

User: Observe this full image without inferring the intended scene. Count actual living people; do not count people represented in paintings, murals, photographs, statues or mirror reflections twice. If the distinction is uncertain, say unknown rather than forcing a count. Describe hands only where visible. Return JSON only:
{"living_person_count":0,"count_status":"observable|unknown","depicted_or_reflected_people":"present|absent|unknown","hands_status":"all_visible_empty|visible_object_held|not_all_hands_visible|unknown","visible_hand_count":0,"object_holding_visible_hand_count":0,"visible_persons_holding_objects":0,"persons_with_observable_hands":0,"hand_count_status":"observable|unknown","held_objects":[],"objects":{"easel":{"presence":"present|absent|unknown","visibility":"clear|partial|not_visible|unknown"},"basketball":{"presence":"present|absent|unknown","visibility":"clear|partial|not_visible|unknown"},"coffee_machine":{"presence":"present|absent|unknown","visibility":"clear|partial|not_visible|unknown"},"watering_can":{"presence":"present|absent|unknown","visibility":"clear|partial|not_visible|unknown"},"building_model":{"presence":"present|absent|unknown","visibility":"clear|partial|not_visible|unknown"},"sculpture_pedestal":{"presence":"present|absent|unknown","visibility":"clear|partial|not_visible|unknown"},"pottery_wheel":{"presence":"present|absent|unknown","visibility":"clear|partial|not_visible|unknown"},"clay_bowl":{"presence":"present|absent|unknown","visibility":"clear|partial|not_visible|unknown"},"bread_rack":{"presence":"present|absent|unknown","visibility":"clear|partial|not_visible|unknown"},"flower_bouquet":{"presence":"present|absent|unknown","visibility":"clear|partial|not_visible|unknown"},"guitar":{"presence":"present|absent|unknown","visibility":"clear|partial|not_visible|unknown"},"musical_score":{"presence":"present|absent|unknown","visibility":"clear|partial|not_visible|unknown"},"chessboard":{"presence":"present|absent|unknown","visibility":"clear|partial|not_visible|unknown"},"camera":{"presence":"present|absent|unknown","visibility":"clear|partial|not_visible|unknown"},"book":{"presence":"present|absent|unknown","visibility":"clear|partial|not_visible|unknown"},"hat":{"presence":"present|absent|unknown","visibility":"clear|partial|not_visible|unknown"},"wig":{"presence":"present|absent|unknown","visibility":"clear|partial|not_visible|unknown"}},"evidence":"brief visible evidence"}.
living_person_count must be a nonnegative integer when observable and null otherwise. held_objects lists actual objects held in visible hands, or [] only when no held object is visible; hands_status must still disclose hidden hands. Count only physically visible hands: visible_hand_count is the number of visible actual-person hands, object_holding_visible_hand_count is the subset visibly holding an object, visible_persons_holding_objects is the number of actual people with at least one visibly object-holding hand, and persons_with_observable_hands is the number of actual people with at least one physically visible hand. A book visibly held with two hands counts as two object-holding hands. Never count hidden hands or infer that hidden hands are empty. Use nonnegative integers only when hand_count_status is observable; otherwise set these four hand-count fields to null and hand_count_status to unknown. The object-holding hand count cannot exceed the visible hand count; object-holding people cannot exceed people with observable hands, and each observable person must contribute at least one visible hand. For each fixed object class use {"presence":"present|absent|unknown","visibility":"clear|partial|not_visible|unknown"}. 'absent' means a sufficiently visible region has no such object; do not infer object absence behind people or outside the image. Observe generic classes consistently; a generic bowl is not necessarily a clay bowl, generic shelves are not necessarily a bread rack, a visible sculpture does not imply a sculpture pedestal.
\end{lstlisting}

\subsection{Count Questions}
Short count question (output limit 240 tokens):
\begin{lstlisting}
Observe this actual full image. Count physically present living people. Do not count statues, sculptures, paintings, murals, photographs or displays as living people. Do not count a mirror reflection as an additional person. A partly visible real person still counts. If visibility prevents a reliable count, report unknown. Do not infer the intended count. Return JSON only with exactly these keys: {"living_person_count":0, "count_status":"observable|unknown", "depicted_or_reflected_people":"present|absent|unknown", "evidence":"brief visible evidence"}. When count_status is unknown, living_person_count must be null. Otherwise it must be a nonnegative integer.
\end{lstlisting}
Count-only question:
\begin{lstlisting}
Inspect the whole frame and enumerate physically distinct living people by their approximate position. Use the exact same visible scene; do not infer the intended number. A statue, bust, portrait, mural, screen image or the reflection of an already counted person is not an additional living person. Include a partly visible person when it is identifiable as physically present. Do not assess clothes, hairstyle, actions or prompt compliance. Return JSON only: {"living_person_count": null, "count_status": "observable|unknown", "person_locations": ["brief location of each counted person"], "evidence": "brief reason for the count or uncertainty"}. If the physical number cannot be established, use unknown and null. Otherwise use a nonnegative integer.
\end{lstlisting}

\subsection{Place and Activity Question}
The option lists are filled with the scenario's place descriptions and activities in a shuffled order (output limit 360 tokens). The landmark answers requested in Question~3 are not used for scoring.
\begin{lstlisting}
Observe this actual full image. Use only visible evidence; do not guess an intended story or setting. Question 1 (place): which ONE option best describes the visible setting? P1: <place> ... P5: <place>; P0: none of these; PX: cannot tell from the image. Question 2 (activity): which ONE option best describes what the visible person or people are doing? A1: <activity> ...; A0: none of these; AX: cannot tell from the image. Question 3 (landmarks): for each item answer present, absent, or unknown. Use unknown when the relevant region is not visible enough to decide. Return JSON only with exactly these keys: {"place":"P1|...|P0|PX", "activity":"A1|...|A0|AX", "landmarks":{...}, "evidence":"brief visible evidence"}
\end{lstlisting}

\section{LLM Instructions}
\label{app:llm_prompts}
All calls share one system instruction: the model is a component of a video-generation system, it sees only the text the system provides, it must not guess appearances, identities, or places that the input does not state, it reports insufficient input in a designated field, and it returns one JSON object.

\paragraph{Record construction (scene commit).}
The input is the scene's content prompt, the entities introduced in earlier scenes, and the detection list of every source (detection ID, detector, label, and box).
The instructions are:
(1) name each person as in the content prompt and reuse the earlier entity ID when the name matches, never linking two people by similar clothing labels;
(2) for every source, pick the person detection of that entity, or mark the source unresolved;
(3) inside that detection, collect the face, the head hair with the face and garment boxes removed, and the union of all garment detections;
(4) describe each component only with the detector labels it uses, without adding colors, lengths, or garments;
(5) list unused detections, such as objects the prompt does not mention, with a reason;
(6) build the background record by excluding every person and component box, and describe it with the place phrase of the prompt and the structures detected in most sources.

\paragraph{Routing.}
The input is the content prompt, the reuse query, the introduction scene of each entity, and the catalog, which lists for every past scene its content prompt and its records with descriptions and availability.
The instructions are:
(1) create one role for each person the query asks to recall and a background role only if it asks for a past place;
(2) choose the version scene that the query points to, by scene number mentioned in the queries and by the place or activity of the scene in the queries, never defaulting to the most recent scene and reporting a tie as ambiguous;
(3) take the face from the person's introduction scene and the head hair and garments from the chosen version;
(4) leave out any component whose attribute the content prompt changes, so that the prompt supplies it;
(5) select only available records and report missing ones without substituting another person, scene, or background.
The output lists, for each role, the entity, the version scene, and the scene of each component, together with the background scene and any missing or ambiguous roles.

\section{Scenario-Authoring Prompt}
\label{app:authoring_prompt}
The complete instruction used to author each scenario, including the motif catalog and the twenty assignment cards, is reproduced below.
\begin{lstlisting}
SELF-CONTAINED SCENARIO-AUTHORING PROMPT v0.5

RUN SETTINGS
TARGET_SCENARIO_ID = MAIN01
OUTPUT_SCENARIOS = 1

Change only TARGET_SCENARIO_ID to select MAIN01 through MAIN20 for a new
  authoring request. Apply the same common instructions to every
  request. Return one complete story, not one scene and not the entire
  twenty-story set. All necessary motifs, assignments and output
  requirements are embedded below. You do not need an external brief or
  an earlier conversation.

ROLE AND PURPOSE
You are authoring a benchmark for selective visual-memory reuse in
  autoregressive text-to-video generation. Write short English scene
  prompts, two equivalent forms of retrieval query, and a structured
  evaluation specification. The current task is authoring benchmark
  inputs, not generating video, routing real KV, or judging a method's
  performance. The assignment cards are a proposed authoring design;
  they are not verified generated observations or a frozen final
  benchmark.

MEMFLOW BORROWING RULE
Use the assigned human-centered role, place and visible-activity motifs
  from the approved catalog. Borrow themes only. Write NEW sentences,
  connected stories, appearance edits and queries. Do not copy original
  scene scripts or claim to reproduce the MemFlow benchmark. Do not
  invent additional MemFlow source rows. Roles, names, gender
  presentation, wardrobe transitions and the branching protocol in the
  cards are our authoring choices, not claims about the original
  narratives.
Pinned source: KlingAIResearch/MemFlow, commit
  7ed514771f5add2af7cc60889e8310ce2b70ab1e,
  prompts/interactive_benchmark.jsonl.
Source URL: https://github.com/KlingAIResearch/MemFlow/blob/7ed514771f5a
  dd2af7cc60889e8310ce2b70ab1e/prompts/int
  eractive_benchmark.jsonl
Source SHA256:
  2f60b571b1b4f29fae6636261fbf61f4a73d81c3
  5406100d73c4287798499938
M003 means source line 3; MAIN03 means one new scenario; S3 means one
  scene. Do not confuse these identifiers.

FIXED STORY AND SOURCE STRUCTURE
Read ONLY the card selected by TARGET_SCENARIO_ID as the current story
  assignment. Use its two people, motifs, assigned appearance states and
  primary edits. Other cards show dataset-level balance; do not merge
  their actors or settings into the output.
S1: Introduce P1 alone at L1, using P1's assigned initial clothing and
  head hair. Use the assigned MemFlow role/place/activity motif.
S2: Introduce the distinct P2 alone at L2, using P2's assigned initial
  appearance and motif. This is a new person, not a recall of P1.
S3: Select P1 from S1, move to NEW L3, and apply exactly P1's assigned
  primary edit. Preserve face identity and all protected appearance
  fields. Exactly one person: only the selected P1; P2 is absent.
S4: Select P2 from S2, move to NEW L4, and apply exactly P2's assigned
  primary edit. Preserve identity and protected fields. Make L4 a
  clearly describable place for later recall. Exactly one person: only
  the selected P2; P1 is absent.
S5: Select P1's S1 appearance and P2's S2 appearance together at NEW L5.
  Exactly two people; empty hands; one simple sustained shared
  interaction.
S6: Start five INDEPENDENT branches from the SAME completed S5 state.
  Write one shared content prompt, not five different content prompts.
  Both selected people keep their selected appearances and selected
  background layout while interacting with empty hands.
S6-A: P1 from S1, P2 from S2, background from S4.
S6-B: P1 from S1, P2 from S4, background from S4.
S6-C: P1 from S3, P2 from S2, background from S4.
S6-D: P1 from S3, P2 from S4, background from S4.
S6-E: P1 from S3, P2 from S4, background from S1.
S6 is not a new L6: its selected place is L4 or L1. Do not continue A
  into B or one branch into another. The branch mapping is immutable.

CONNECTED NARRATIVE
Use the supplied story_link as a light connection between the people's
  activities and later meeting. Keep the plot simple. Each scene depicts
  one setting and one simple sustained action or stable state. Do not
  require exact one-off completion, readable text, elaborate finger
  movements, crowds, rapid cuts or multiple action chains. Borrow only
  the motif; simplify source activities that would hide the face or need
  tiny manipulation. Do not introduce unassigned extra people, children,
  animals or supernatural transformations.
Use five distinct settings L1-L5 with modest, observable structure. L1
  and L4 each need two or three large landmarks or layout relations for
  background recall. Do not rely on a place-category word alone. Never
  include dialogue, emotional subplots or sound requirements needed to
  understand the task.

APPEARANCE AND EDIT RULES
Treat identity and editable appearance separately. Maintain the same
  face identity, adult age and body build through S1/S3 and S2/S4. Write
  an initial face description that distinguishes the two people without
  claiming the video model will realize it.
The card's intended_initial and intended_edited refer to clothing for
  clothing edits and to head hair for hair edits. Use initial_clothing
  and initial_head_hair to complete each initial appearance. For hair
  edits keep clothing_to_preserve unchanged.
Color edit: change only the broad color of the specified garment;
  preserve its category, sleeve length, layers and hairstyle.
Category edit: change only the assigned garment category; preserve broad
  color and sleeve length. Preserve an explicitly retained undershirt.
  Keep all hoods down.
Layer edit: add/remove the assigned apron, vest or cardigan; preserve
  the underlying garment's category, color and sleeves. Newly visible
  underlying clothing is not an additional requested edit.
Sleeve edit: change long versus short sleeves; preserve garment
  color/category, other layers and hair. Make both relevant arms
  visible.
Hair edits: distinguish head-hair presence, length, texture and
  arrangement. Long curly hair can become long straight hair at fixed
  length/color; long straight hair can become short straight hair at
  fixed texture/color; short natural hair can become short slicked-back
  hair with a visible forehead at fixed length/color. The bald
  transition changes hair presence; length, texture, arrangement and
  color become not applicable when head hair is absent. Keep face,
  apparent age, body build, clothing and facial hair unchanged. A bald
  scalp must be visible, not hidden by a hat, crop or wig.
Do not request glasses or facial-hair edits. Do not request hair-color,
  age, gender, facial-shape or body-build changes. Avoid decorative
  accessories not assigned by the card. Describe the edited state
  ALREADY present in the new scene, not an on-screen haircut, clothes
  change, morph or hair-growth event.

PROMPT LENGTH AND VISIBILITY
Write concrete, grammatical, short English sentences. Match the length
  of MemFlow scene prompts: target 30-36 whitespace-separated words per
  content prompt, with a HARD MAXIMUM of 40 words including camera
  instructions. The pinned source has 600 scene prompts: mean 33.04
  words, median 33, and 90th percentile 39. Remove decorative prose and
  redundant camera adjectives first. Preserve the visible person count,
  location/action, requested edit, protected identity intent, and
  required head/arm visibility. Do not move necessary visible
  requirements into evaluator-only metadata. Keep retrieval queries
  concise too; they are routing inputs, not extra appearance
  descriptions. If an adapter also supplies query text to the video
  model, record its total actual conditioning length separately rather
  than claiming that only content_prompt was supplied.
Use a fixed medium-wide view with faces and relevant appearance visible.
  Show full head hair for hair judgments and arms for sleeves. Keep
  books/cameras below or away from faces. Any forbidden held prop must
  have been relevant to that story and stated in the current content
  prompt if it is to be evaluated. Source scenes may contain their
  natural props. S5/S6 require empty hands, but normal furniture and
  background fixtures remain allowed.

CONTENT p VERSUS RETRIEVAL q
Keep the content prompt and source-selection query in separate fields.
  S3/S4 content states the requested NEW attribute state. It must not
  re-describe an unedited face as a text substitute for identity memory.
S5 and S6_shared content refers to selected people/appearances and the
  selected place. DO NOT repeat the answer garment colors, garment
  categories, hairstyles, source scene numbers, source places or source
  role descriptions there. All five S6 branches share the exact same
  content string.
Provide explicit and natural queries for S3, S4, S5 and S6 A-E. Explicit
  queries use actor names and scene numbers. Natural queries use names
  and past place/activity cues without scene numbers or answer
  attributes. Each pair must identify exactly the same intended source
  roles. Example forms: 'Use [name] from Scene 1' and 'Use [name] as
  they appeared while [past activity] in [past place]'. Do not copy this
  example literally. If a cue is ambiguous, report it instead of
  silently selecting a different source. For background queries,
  identify the source by its place and the actor's past activity. Do not
  repeat the target background's structural landmarks, materials, or
  spatial layout as an answer description when the source is already
  uniquely identifiable. Apply the same rule to actor-version cues:
  avoid unnecessary landmark details.

INTENT VERSUS OBSERVATION
All appearance descriptions here are intended states. Do not write
  observed_state, successful_identity, generated_result or performance
  claims. A later evaluator must independently annotate actual RGB,
  source identity continuity, state observability and version
  differences. Attribute values and face identity must be bound to the
  same target actor. Future branch answers and evaluator annotations
  must not be input to the generator's earlier memory formation or
  retention.

OUTPUT CONTRACT
Return exactly ONE JSON object, with these top-level keys and no
  Markdown fences:
1. scenario_id: the selected MAINxx ID.
2. motif_provenance: two objects containing actor_id, motif_id,
  source_line_1based, source_commit and the pinned source URL with its
  line anchor.
3. story_summary: one short paragraph connecting the scenes.
4. actors: P1 and P2, each containing name, adult_gender_presentation,
  initial_face_description, intended_initial_appearance and
  protected_attributes. Appearance uses nested clothing and head_hair
  fields; use null for inapplicable hair properties, never an invented
  observed value.
5. attribute_transitions: P1 and P2, each containing from_scene,
  to_scene, primary_edit_family, primary_attribute_path,
  intended_before, intended_after, protected_attributes,
  required_visibility and stress_tags. Preserve the assigned edit; do
  not substitute another.
6. backgrounds: L1-L5, each containing introduced_in_scene,
  place_description and structural_cues. Use two or three large cues for
  L1/L4 and mark them as intended.
7. scene_templates: exactly S1, S2, S3, S4, S5 and S6_shared. Each
  contains content_prompt and word_count. Do not write five
  branch-specific S6 content prompts.
8. retrieval_queries: S3, S4 and S5 each containing explicit and natural
  strings; S6 containing A-E, each with explicit and natural strings.
  S1/S2 require no reuse query.
9. expected_source_roles: S3={P1:S1}; S4={P2:S2}; S5={P1:S1,P2:S2,bg:new
  L5}; and the exact S6 A-E map given above. These are evaluation-intent
  labels, not real KV record IDs.
10. evaluation_manifest: per-scene count, action, place requirements,
  intended edited and protected attributes, required visible regions and
  explicitly requested exclusions. S6 must include actor-specific
  selected states and selected background as evaluator-only intent, not
  content text.
11. authoring_checks: word counts; exact-branch-map check; actor-lineage
  check; one-primary-edit check; protected-field consistency; query-pair
  equivalence; visibility; no appearance-answer leakage; and
  unresolved_constraints. Checks are text/schema checks, not proof of
  video success.

QUALITY CONTROL AND REPRODUCIBILITY
Apply the same rules to all cards. Do not choose easier materials, add
  appearance answers, or select stories based on any method's generated
  videos. If an assignment cannot satisfy a rule, keep the specified
  assignment and place the reason in unresolved_constraints; do not
  silently replace it. Do not fabricate model version, sampling
  settings, request IDs or validation receipts. Those are recorded
  externally along with raw responses, revisions and the final frozen
  JSON.

APPROVED MEMFLOW MOTIF CATALOG
Columns: ID | role motif | place motif | adapted simple activity |
  authoring caution.
M003 | chef | restaurant kitchen | plating food | Simplify tiny food
  manipulation; keep face visible.
M004 | street artist | mural-lined alley | painting a wall | Use one
  adult; omit other source characters.
M012 | reader | park bench | reading a book | Keep the book below the
  face.
M013 | guitarist | street corner | playing guitar | Do not score exact
  finger motion.
M017 | teacher | classroom | explaining beside a chalkboard | Use one
  adult; omit pupils and readable equations.
M019 | programmer | desk in a workroom | typing | Avoid dark lighting
  and facial backlighting.
M021 | photographer | fountain plaza | taking photographs | Avoid
  continuous camera occlusion of the face.
M024 | park ranger | wooded trail | walking | Omit animal-rescue
  interactions.
M027 | chess player | wood-paneled hall | studying a chessboard | Do not
  require legal chess moves.
M034 | vineyard worker | vineyard | inspecting grapes | Avoid fine
  grape-by-grape manipulation.
M037 | camper | pine-forest clearing | tidying beside a tent | Do not
  require full tent assembly.
M044 | pianist | empty concert hall | playing piano | Keep face and
  relevant upper-body attributes visible.
M047 | writer | book-lined attic | typing at a typewriter | No readable
  generated text.
M050 | postal worker | suburban street | sorting mail beside a mailbox |
  Use a sustained simple action; omit animal interactions.
M058 | archaeologist | excavation site | brushing soil | Avoid requiring
  precise fossil restoration.
M060 | street cleaner | quiet city street | sweeping | Use lighting
  sufficient for identity and attribute observation.
M064 | conductor | concert hall | studying a score | Omit an orchestra
  crowd and readable music notation.
M075 | meditator | garden | seated meditation | Keep face and edited
  attributes visible.
M090 | diary writer | window-side cafe | writing in a notebook | Avoid
  persistent face occlusion and readable text.
M100 | scientist | laboratory | examining equipment beside a microscope
  | Avoid persistent eye-to-eyepiece face occlusion.

PRESPECIFIED SCENARIO CARDS
Use only the card with scenario_id matching TARGET_SCENARIO_ID.
  [{"scenario_id":"MAIN01","actors":{"P1":
  {"name":"Leo","adult_gender_pres
  entation":"man","motif_id":"M003","prima
  ry_edit_family":"clothing_laye
  r_presence","primary_edit_subtype":"laye
  r","intended_initial":"beige
  apron over a white long-sleeved shirt","intended_edited":"white
  long-sleeved shirt without the apron","initial_clothing":"beige apron
  over a white long-sleeved shirt","initial_head_hair":"long curly brown
  hair"},"P2":{"name":"Nora","adult_gender
  _presentation":"woman","motif_
  id":"M044","primary_edit_family":"garmen
  t_color","primary_edit_subtype
  ":"color","intended_initial":"red long-sleeved
  shirt","intended_edited":"blue long-sleeved
  shirt","initial_clothing":"red long-sleeved
  shirt","initial_head_hair":"long straight brown
  hair"}},"story_link":"Preparing food and music for a neighborhood even
  t."},{"scenario_id":"MAIN02","actors":{"
  P1":{"name":"Rafael","adult_ge
  nder_presentation":"man","motif_id":"M00
  4","primary_edit_family":"garm
  ent_category","primary_edit_subtype":"ty
  pe","intended_initial":"blue
  long-sleeved shirt","intended_edited":"blue long-sleeved
  sweater","initial_clothing":"blue long-sleeved
  shirt","initial_head_hair":"short straight brown hair in a natural arr
  angement"},"P2":{"name":"Iris","adult_ge
  nder_presentation":"woman","mo
  tif_id":"M021","primary_edit_family":"cl
  othing_layer_presence","primar
  y_edit_subtype":"layer","intended_initial":"beige vest over a navy
  shirt","intended_edited":"navy shirt without the
  vest","initial_clothing":"beige vest over a navy
  shirt","initial_head_hair":"fully bald scalp"}},"story_link":"Making
  and documenting a community mural."},{"scenario_id":"MAIN03","actors":
  {"P1":{"name":"Elena","adult_gender_pres
  entation":"woman","motif_id":"
  M012","primary_edit_family":"garment_col
  or","primary_edit_subtype":"co
  lor","intended_initial":"yellow long-sleeved
  shirt","intended_edited":"purple long-sleeved
  shirt","initial_clothing":"yellow long-sleeved
  shirt","initial_head_hair":"short brown hair slicked back with a
  visible forehead"},"P2":{"name":"Olivia","adult_gender_presentation":"
  woman","motif_id":"M047","primary_edit_f
  amily":"garment_category","pri
  mary_edit_subtype":"type","intended_initial":"green long-sleeved
  sweater","intended_edited":"green long-sleeved hoodie with the hood
  down","initial_clothing":"green long-sleeved
  sweater","initial_head_hair":"long curly brown
  hair"}},"story_link":"Preparing to meet at a local literary event."},{
  "scenario_id":"MAIN04","actors":{"P1":{"
  name":"Julian","adult_gender_p
  resentation":"man","motif_id":"M013","pr
  imary_edit_family":"head_hair"
  ,"primary_edit_subtype":"hair_texture","intended_initial":"long curly
  brown hair","intended_edited":"long straight brown
  hair","initial_clothing":"navy long-sleeved
  shirt","initial_head_hair":"long curly brown
  hair","clothing_to_preserve":"navy long-sleeved shirt"},"P2":{"name":"
  Colin","adult_gender_presentation":"man"
  ,"motif_id":"M064","primary_ed
  it_family":"garment_color","primary_edit
  _subtype":"color","intended_in
  itial":"white long-sleeved shirt","intended_edited":"blue long-sleeved
  shirt","initial_clothing":"white long-sleeved
  shirt","initial_head_hair":"short straight brown hair in a natural
  arrangement"}},"story_link":"Preparing separate parts of a small music
  performance."},{"scenario_id":"MAIN05","
  actors":{"P1":{"name":"Priya",
  "adult_gender_presentation":"woman","mot
  if_id":"M017","primary_edit_fa
  mily":"sleeve_length","primary_edit_subt
  ype":"sleeve","intended_initia
  l":"red long-sleeved shirt","intended_edited":"red short-sleeved
  shirt","initial_clothing":"red long-sleeved
  shirt","initial_head_hair":"fully bald scalp"},"P2":{"name":"Elise","a
  dult_gender_presentation":"woman","motif
  _id":"M100","primary_edit_fami
  ly":"garment_category","primary_edit_sub
  type":"type","intended_initial
  ":"blue long-sleeved jacket over a white
  shirt","intended_edited":"blue long-sleeved cardigan over the same
  white shirt","initial_clothing":"blue long-sleeved jacket over a white
  shirt","initial_head_hair":"short brown hair slicked back with a
  visible forehead"}},"story_link":"Preparing contributions to a science
  fair."},{"scenario_id":"MAIN06","actors"
  :{"P1":{"name":"Minho","adult_
  gender_presentation":"man","motif_id":"M
  019","primary_edit_family":"ga
  rment_color","primary_edit_subtype":"col
  or","intended_initial":"green
  short-sleeved T-shirt","intended_edited":"red short-sleeved
  T-shirt","initial_clothing":"green short-sleeved
  T-shirt","initial_head_hair":"long curly brown hair"},"P2":{"name":"Ad
  am","adult_gender_presentation":"man","m
  otif_id":"M027","primary_edit_
  family":"clothing_layer_presence","prima
  ry_edit_subtype":"layer","inte
  nded_initial":"gray cardigan over a white
  shirt","intended_edited":"white shirt without the
  cardigan","initial_clothing":"gray cardigan over a white
  shirt","initial_head_hair":"long straight brown
  hair"}},"story_link":"Preparing a demonstration connecting computing
  and board games."},{"scenario_id":"MAIN07","actors":{"P1":{"name":"Gab
  riel","adult_gender_presentation":"man",
  "motif_id":"M024","primary_edi
  t_family":"garment_category","primary_ed
  it_subtype":"type","intended_i
  nitial":"yellow long-sleeved sweater","intended_edited":"yellow
  long-sleeved hoodie with the hood down","initial_clothing":"yellow
  long-sleeved sweater","initial_head_hair":"short straight brown hair
  in a natural arrangement"},"P2":{"name":"Lina","adult_gender_presentat
  ion":"woman","motif_id":"M058","primary_
  edit_family":"head_hair","prim
  ary_edit_subtype":"hair_length","intended_initial":"long straight
  brown hair","intended_edited":"short straight brown
  hair","initial_clothing":"green long-sleeved
  shirt","initial_head_hair":"long straight brown
  hair","clothing_to_preserve":"green long-sleeved
  shirt"}},"story_link":"Preparing nature and heritage information for a
  trail exhibition."},{"scenario_id":"MAIN08","actors":{"P1":{"name":"Hu
  go","adult_gender_presentation":"man","m
  otif_id":"M034","primary_edit_
  family":"garment_color","primary_edit_su
  btype":"color","intended_initi
  al":"blue short-sleeved polo shirt","intended_edited":"green
  short-sleeved polo shirt","initial_clothing":"blue short-sleeved polo
  shirt","initial_head_hair":"short brown hair slicked back with a
  visible forehead"},"P2":{"name":"Mira","adult_gender_presentation":"wo
  man","motif_id":"M037","primary_edit_fam
  ily":"clothing_layer_presence"
  ,"primary_edit_subtype":"layer","intended_initial":"navy short-sleeved
  T-shirt without a vest","intended_edited":"beige vest over the same
  navy short-sleeved T-shirt","initial_clothing":"navy short-sleeved
  T-shirt without a vest","initial_head_hair":"long curly brown
  hair"}},"story_link":"Preparing for a countryside harvest gathering
  near a campsite."},{"scenario_id":"MAIN09","actors":{"P1":{"name":"Jon
  as","adult_gender_presentation":"man","m
  otif_id":"M050","primary_edit_
  family":"clothing_layer_presence","prima
  ry_edit_subtype":"layer","inte
  nded_initial":"yellow vest over a gray long-sleeved
  shirt","intended_edited":"gray long-sleeved shirt without the
  vest","initial_clothing":"yellow vest over a gray long-sleeved
  shirt","initial_head_hair":"long straight brown hair"},"P2":{"name":"N
  ina","adult_gender_presentation":"woman"
  ,"motif_id":"M060","primary_ed
  it_family":"garment_category","primary_e
  dit_subtype":"type","intended_
  initial":"green short-sleeved T-shirt","intended_edited":"green
  short-sleeved polo shirt","initial_clothing":"green short-sleeved
  T-shirt","initial_head_hair":"short straight brown hair in a natural
  arrangement"}},"story_link":"Finishing neighborhood work before a
  community gathering."},{"scenario_id":"MAIN10","actors":{"P1":{"name":
  "Asha","adult_gender_presentation":"woma
  n","motif_id":"M075","primary_
  edit_family":"garment_category","primary
  _edit_subtype":"type","intende
  d_initial":"blue short-sleeved T-shirt","intended_edited":"blue
  short-sleeved polo shirt","initial_clothing":"blue short-sleeved
  T-shirt","initial_head_hair":"fully bald scalp"},"P2":{"name":"Freya",
  "adult_gender_presentation":"woman","mot
  if_id":"M090","primary_edit_fa
  mily":"sleeve_length","primary_edit_subt
  ype":"sleeve","intended_initia
  l":"purple long-sleeved shirt","intended_edited":"purple short-sleeved
  shirt","initial_clothing":"purple long-sleeved
  shirt","initial_head_hair":"short brown hair slicked back with a
  visible forehead"}},"story_link":"Preparing reflection activities for
  a quiet retreat."},{"scenario_id":"MAIN11","actors":{"P1":{"name":"Mar
  co","adult_gender_presentation":"man","m
  otif_id":"M003","primary_edit_
  family":"garment_category","primary_edit
  _subtype":"type","intended_ini
  tial":"brown long-sleeved shirt","intended_edited":"brown long-sleeved
  sweater","initial_clothing":"brown long-sleeved
  shirt","initial_head_hair":"long curly brown hair"},"P2":{"name":"Luca
  ","adult_gender_presentation":"man","mot
  if_id":"M034","primary_edit_fa
  mily":"clothing_layer_presence","primary
  _edit_subtype":"layer","intend
  ed_initial":"white short-sleeved shirt without a
  vest","intended_edited":"gray vest over the same white short-sleeved
  shirt","initial_clothing":"white short-sleeved shirt without a
  vest","initial_head_hair":"long straight brown
  hair"}},"story_link":"Preparing different parts of a harvest supper."}
  ,{"scenario_id":"MAIN12","actors":{"P1":
  {"name":"Soren","adult_gender_
  presentation":"man","motif_id":"M044","p
  rimary_edit_family":"garment_c
  olor","primary_edit_subtype":"color","in
  tended_initial":"purple
  long-sleeved shirt","intended_edited":"yellow long-sleeved
  shirt","initial_clothing":"purple long-sleeved
  shirt","initial_head_hair":"short straight brown hair in a natural arr
  angement"},"P2":{"name":"Yuna","adult_ge
  nder_presentation":"woman","mo
  tif_id":"M013","primary_edit_family":"ga
  rment_category","primary_edit_
  subtype":"type","intended_initial":"red long-sleeved
  sweater","intended_edited":"red long-sleeved hoodie with the hood
  down","initial_clothing":"red long-sleeved
  sweater","initial_head_hair":"fully bald
  scalp"}},"story_link":"Preparing separately for an outdoor concert."},
  {"scenario_id":"MAIN13","actors":{"P1":{
  "name":"Celia","adult_gender_p
  resentation":"woman","motif_id":"M004","
  primary_edit_family":"sleeve_l
  ength","primary_edit_subtype":"sleeve","
  intended_initial":"blue
  short-sleeved shirt","intended_edited":"blue long-sleeved
  shirt","initial_clothing":"blue short-sleeved
  shirt","initial_head_hair":"short brown hair slicked back with a
  visible forehead"},"P2":{"name":"Dario","adult_gender_presentation":"m
  an","motif_id":"M060","primary_edit_fami
  ly":"garment_color","primary_e
  dit_subtype":"color","intended_initial":"red long-sleeved
  shirt","intended_edited":"blue long-sleeved
  shirt","initial_clothing":"red long-sleeved
  shirt","initial_head_hair":"long curly brown
  hair"}},"story_link":"Preparing a renewed public space for a
  neighborhood event."},{"scenario_id":"MAIN14","actors":{"P1":{"name":"
  Victor","adult_gender_presentation":"man
  ","motif_id":"M021","primary_e
  dit_family":"head_hair","primary_edit_su
  btype":"hair_arrangement","int
  ended_initial":"short brown hair in a natural
  arrangement","intended_edited":"short brown hair slicked back with a
  visible forehead","initial_clothing":"red short-sleeved
  shirt","initial_head_hair":"short brown hair in a natural
  arrangement","clothing_to_preserve":"red short-sleeved shirt"},"P2":{"
  name":"Kieran","adult_gender_presentatio
  n":"man","motif_id":"M037","pr
  imary_edit_family":"garment_category","p
  rimary_edit_subtype":"type","i
  ntended_initial":"green long-sleeved jacket over a white
  shirt","intended_edited":"green long-sleeved cardigan over the same
  white shirt","initial_clothing":"green long-sleeved jacket over a
  white shirt","initial_head_hair":"short straight brown hair in a
  natural arrangement"}},"story_link":"Preparing a nature-photography
  outing and discussing the trip."},{"scenario_id":"MAIN15","actors":{"P
  1":{"name":"Hana","adult_gender_presenta
  tion":"woman","motif_id":"M012
  ","primary_edit_family":"garment_color",
  "primary_edit_subtype":"color"
  ,"intended_initial":"red long-sleeved sweater","intended_edited":"blue
  long-sleeved sweater","initial_clothing":"red long-sleeved
  sweater","initial_head_hair":"fully bald scalp"},"P2":{"name":"Sienna"
  ,"adult_gender_presentation":"woman","mo
  tif_id":"M017","primary_edit_f
  amily":"clothing_layer_presence","primar
  y_edit_subtype":"layer","inten
  ded_initial":"white long-sleeved shirt without a
  cardigan","intended_edited":"navy cardigan over the same white
  long-sleeved shirt","initial_clothing":"white long-sleeved shirt
  without a cardigan","initial_head_hair":"short brown hair slicked back
  with a visible forehead"}},"story_link":"Preparing a reading activity
  for a library event."},{"scenario_id":"MAIN16","actors":{"P1":{"name":
  "Ren","adult_gender_presentation":"man",
  "motif_id":"M047","primary_edi
  t_family":"garment_category","primary_ed
  it_subtype":"type","intended_i
  nitial":"yellow short-sleeved T-shirt","intended_edited":"yellow
  short-sleeved polo shirt","initial_clothing":"yellow short-sleeved
  T-shirt","initial_head_hair":"long curly brown hair"},"P2":{"name":"Al
  ma","adult_gender_presentation":"woman",
  "motif_id":"M090","primary_edi
  t_family":"garment_color","primary_edit_
  subtype":"color","intended_ini
  tial":"green long-sleeved shirt","intended_edited":"purple
  long-sleeved shirt","initial_clothing":"green long-sleeved
  shirt","initial_head_hair":"long straight brown
  hair"}},"story_link":"Preparing pieces to discuss at a small writing m
  eetup."},{"scenario_id":"MAIN17","actors
  ":{"P1":{"name":"Peter","adult
  _gender_presentation":"man","motif_id":"
  M064","primary_edit_family":"g
  arment_color","primary_edit_subtype":"co
  lor","intended_initial":"black
  long-sleeved shirt","intended_edited":"white long-sleeved
  shirt","initial_clothing":"black long-sleeved
  shirt","initial_head_hair":"short straight brown hair in a natural arr
  angement"},"P2":{"name":"Leon","adult_ge
  nder_presentation":"man","moti
  f_id":"M019","primary_edit_family":"garm
  ent_category","primary_edit_su
  btype":"type","intended_initial":"purple long-sleeved
  shirt","intended_edited":"purple long-sleeved
  sweater","initial_clothing":"purple long-sleeved
  shirt","initial_head_hair":"fully bald
  scalp"}},"story_link":"Preparing musical and technical contributions
  to a performance."},{"scenario_id":"MAIN18","actors":{"P1":{"name":"An
  ika","adult_gender_presentation":"woman"
  ,"motif_id":"M100","primary_ed
  it_family":"sleeve_length","primary_edit
  _subtype":"sleeve","intended_i
  nitial":"green short-sleeved shirt","intended_edited":"green
  long-sleeved shirt","initial_clothing":"green short-sleeved
  shirt","initial_head_hair":"short brown hair slicked back with a
  visible forehead"},"P2":{"name":"Marta","adult_gender_presentation":"w
  oman","motif_id":"M058","primary_edit_fa
  mily":"garment_color","primary
  _edit_subtype":"color","intended_initial":"blue short-sleeved
  T-shirt","intended_edited":"red short-sleeved
  T-shirt","initial_clothing":"blue short-sleeved
  T-shirt","initial_head_hair":"long curly brown
  hair"}},"story_link":"Preparing scientific and historical displays for
  a museum event."},{"scenario_id":"MAIN19","actors":{"P1":{"name":"Noah
  ","adult_gender_presentation":"man","mot
  if_id":"M027","primary_edit_fa
  mily":"head_hair","primary_edit_subtype"
  :"hair_presence","intended_ini
  tial":"short brown hair","intended_edited":"fully bald scalp without a
  hat or wig","initial_clothing":"blue long-sleeved
  sweater","initial_head_hair":"short brown
  hair","clothing_to_preserve":"blue long-sleeved sweater","stress_tag":
  "large_head_hair_change"},"P2":{"name":"
  Mei","adult_gender_presentatio
  n":"woman","motif_id":"M075","primary_ed
  it_family":"garment_color","pr
  imary_edit_subtype":"color","intended_initial":"dark blue long-sleeved
  sweater","intended_edited":"yellow long-sleeved
  sweater","initial_clothing":"dark blue long-sleeved
  sweater","initial_head_hair":"short straight brown hair in a natural
  arrangement"}},"story_link":"Preparing concentration activities for a
  community event."},{"scenario_id":"MAIN20","actors":{"P1":{"name":"Idr
  is","adult_gender_presentation":"man","m
  otif_id":"M024","primary_edit_
  family":"clothing_layer_presence","prima
  ry_edit_subtype":"layer","inte
  nded_initial":"green vest over a beige long-sleeved
  shirt","intended_edited":"beige long-sleeved shirt without the
  vest","initial_clothing":"green vest over a beige long-sleeved
  shirt","initial_head_hair":"fully bald scalp"},"P2":{"name":"Eva","adu
  lt_gender_presentation":"woman","motif_i
  d":"M050","primary_edit_family
  ":"garment_category","primary_edit_subty
  pe":"type","intended_initial":
  "yellow long-sleeved sweater","intended_edited":"yellow long-sleeved
  hoodie with the hood down","initial_clothing":"yellow long-sleeved
  sweater","initial_head_hair":"short brown hair slicked back with a
  visible forehead"}},"story_link":"Preparing to meet at a rural
  community gathering."}]

Now produce the one requested JSON object.
\end{lstlisting}

\end{document}